\documentclass[10pt,journal,cspaper,compsoc,twoside]{IEEEtran}
\pdfoutput=1
\usepackage[breaklinks=true,bookmarks=false]{hyperref}
\usepackage{graphicx}
\usepackage{amsmath}
\usepackage{amssymb}
\usepackage{url}
\usepackage{rotating}
\usepackage{multirow}
\usepackage{wrapfig}
\usepackage{tabularx}
\usepackage{color}

\newcommand*\Bell{\ensuremath{\boldsymbol\ell}}
\newcommand*\bs{\ensuremath{\boldsymbol}}
\newcommand{\rmbf}[1]{\ensuremath{\mathrm{\bf #1}}}

\ifCLASSOPTIONcompsoc
  \usepackage[nocompress]{cite}
\else
  \usepackage{cite}
\fi
\ifCLASSINFOpdf
\else
\fi
\ifCLASSOPTIONcompsoc
\usepackage[tight,normalsize,sf,SF]{subfigure}
\else
\usepackage[tight,footnotesize]{subfigure}
\fi
\begin{document}
\title{
    Object Detection Through Exploration With A Foveated Visual Field \\
}

\author{Emre Akbas, Miguel P. Eckstein
        %John~Doe,~\IEEEmembership{Fellow,~OSA,}
        %and~Jane~Doe,~\IEEEmembership{Life~Fellow,~IEEE}% <-this % stops a space
\IEEEcompsocitemizethanks{\IEEEcompsocthanksitem E. Akbas and M. P. Eckstein are
    with the Vision and Image Understanding Laboratory at the Department of
Psychological and Brain Sciences, University of California Santa Barbara. }
%of Electrical and Computer Engineering, Georgia Institute of Technology, Atlanta,
%GA, 30332.\protect\\
%% note need leading \protect in front of \\ to get a newline within \thanks as
%% \\ is fragile and will error, could use \hfil\break instead.
%E-mail: see http://www.michaelshell.org/contact.html
%\IEEEcompsocthanksitem J. Doe and J. Doe are with Anonymous University.}% <-this % stops a space
\thanks{}}


\markboth{}
{
    Object Detection Through Exploration With A Foveated Visual Field 
}

%\IEEEpubid{\makebox[\columnwidth]{\hfill 0000--0000/00/\$00.00~\copyright~2007 IEEE}%

% PRIOR to the title within the \IEEEcompsoctitleabstractindextext IEEEtran
\IEEEcompsoctitleabstractindextext{%
\begin{abstract}
    We present a foveated object detector (FOD) as a biologically-inspired
    alternative to the sliding window (SW) approach which is the dominant method
    of search in computer vision object detection. Similar to the human visual
    system, the FOD has higher resolution at the fovea and lower resolution at
    the visual periphery. Consequently, more computational resources are
    allocated at the fovea and relatively fewer at the periphery. The FOD
    processes the entire scene, uses retino-specific object detection
    classifiers to guide eye movements, aligns its fovea with regions of
    interest in the input image and integrates observations across multiple
    fixations. Our approach combines modern object detectors from computer
    vision with a recent model of peripheral pooling regions found at the V1
    layer of the human visual system. We assessed various eye movement
    strategies on the PASCAL VOC 2007 dataset and show that the FOD performs on
    par with the SW detector while bringing significant computational cost
    savings. 
\end{abstract}
% not want either math or citations to appear in the abstract.

% Note that keywords are not normally used for peer review papers.
}

\maketitle

% is not selected <OR> if conference mode is selected - because compsoc
% conference papers position the abstract like regular (non-compsoc)
\IEEEdisplaynotcompsoctitleabstractindextext

% \fi
\IEEEpeerreviewmaketitle

% #########################################################################
\section{Introduction}

There has been substantial progress (e.g. \cite{dpm:pami2010, zhu:cvpr2010,
malisiewicz:iccv2011, reng:cvpr2013, vandeSande:iccv2011, kontschieder:nips2012,
dean:cvpr2013} to name a few) in object detection research in recent years.
However, humans are
still unsurpassed in their ability to search
for objects in visual scenes. The human brain relies on a variety of strategies
\cite{Eckstein2011}  including prior
probabilities of object occurrence, global scene statistics \cite{Torralba2006,
Neider2006}  and object co-occurrence  \cite{Eckstein2006,Mack2011,Preston2013}
to successfully detect objects in cluttered scenes.  Object
detection approaches have increasingly included some of the human
strategies \cite{alexe:nips2012,elder:ijcv2007,bazzani:icml2011, dpm:pami2010,
serre:cvpr2005}.
One remaining crucial difference between the human visual system and a modern
object detector is that while humans process the visual field with decreasing
resolution away  \cite{Wertheim1894, Levi1985, Rovamo1984, Strasburger2011}
from the
fixation point and
make saccades to collect information, typical object detectors
\cite{dpm:pami2010}
scan all locations at the same resolution and repeats this at multiple scales.
The goal of the present work is to investigate the impact on object detector
performance of using a foveated visual field and saccade exploration rather than
the dominant sliding window paradigm \cite{dpm:pami2010, zhu:cvpr2010,
malisiewicz:iccv2011}. Such endeavor is of interest
for two reasons.  . First, from the computer vision perspective, using a visual
field with varying resolution might lead to reduction in computational
complexity, consequently the approach might lead to more efficient object
detection algorithms. Second, from a scientific perspective, if a foveated
object detection model can achieve similar performance accuracy as a
non-foveated sliding window approach, it might suggest a possible reason for
the evolution of foveated systems in organisms: achieving successful object
detection while minimizing computational and metabolic costs. 

Contemporary object detection research can be roughly outlined by the following three
important components of a modern object detector: the features, the detection
model and the search model. The most popular choices for
these three components are Histogram of Oriented Gradients (HOG) features
\cite{dalal_triggs:cvpr2005}, mixture of linear templates \cite{dpm:pami2010},
and the sliding window (SW) method, respectively. Although there are efforts to go
beyond these standard choices (e.g. new features \cite{reng:cvpr2013,
vandeSande:iccv2011}; alternative detection models \cite{kontschieder:nips2012,
vandeSande:iccv2011}, whether object parts should be modeled or not
\cite{zhu:bmvc2012,divvala:eccvw2012}; and  alternative search methods
\cite{felzenswalb:cvpr2010, lampert:pami2009,dpm:pami2010, kokkinos:nips2011,
vandeSande:iccv2011}), HOG, mixture of linear templates and SW form the crux
of modern object detection methods (\cite{dpm:pami2010, zhu:cvpr2010,
malisiewicz:iccv2011}).  Here, we build upon the ``HOG + mixture of
linear templates" framework and propose a biologically inspired alternative
search model to the sliding window method, where the detector
searches for the object by making saccades instead of processing all locations
at fine spatial resolution (See Section \ref{sec:related_work} for a more
detailed discussion on related work). 

\begin{figure}
    \centering
    \includegraphics[scale=.5]{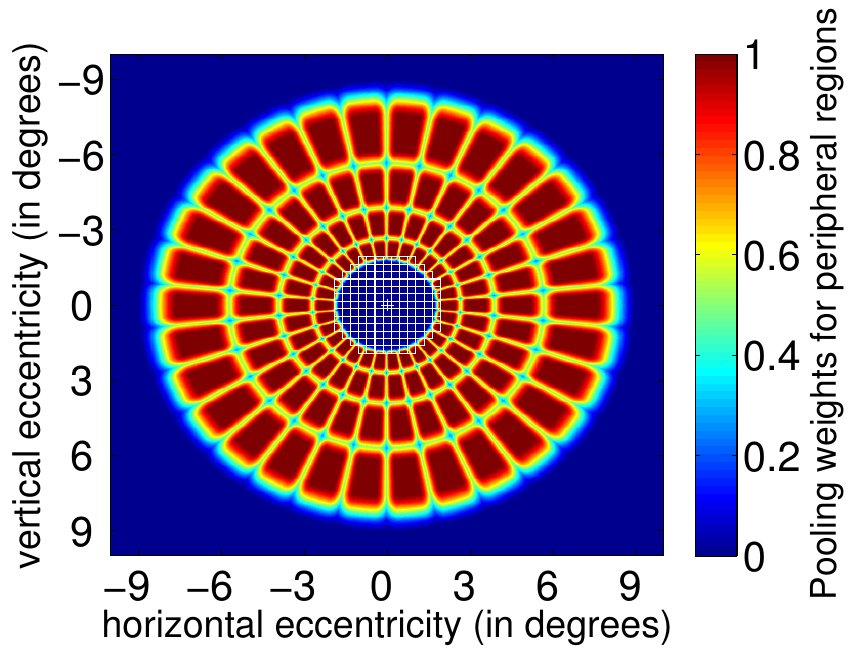} 
    \caption{
        \label{fig:teaser_foveated_model}
        The foveated visual field of the proposed object detector. Square blue boxes
    with white borders at the center are foveal pooling  regions. Around them
    are peripheral pooling regions which are radially elongated. The sizes of
    peripheral regions increase with distance to the fixation point which is at
    the center of the fovea. The color within the peripheral regions represent
    pooling weights. 
}
\end{figure}

The human visual system is known to have a varying resolution visual field. The
fovea has higher resolution and this resolution decreases towards the periphery
\cite{Wertheim1894, Levi1985, Rovamo1984, Strasburger2011}.
As a
consequence, the visual input at and around the fixation location has more
detail relative to peripheral locations away from the fixation point.  Humans
and other mammals make saccades to  align their high resolution fovea with the
regions of interest in the visual environment. There are many possible methods
to
implement such a foveated visual field in an object detection system. In this
work,
we opt to use a recent model \cite{freeman_simoncelli:metamers} which specifies
how responses of elementary sensors are pooled at the layers (V1 and V2) of the human visual cortex. The
model specifies the shapes and sizes of V1, V2 regions which pool responses from
the visual field.
We use a simplified version of this model as the
foveated visual field of our object detector (Figure
\ref{fig:teaser_foveated_model}).  We call our detector as ``the
foveated object detector (FOD)" due to its foveated visual field. 

The sizes of pooling regions in the visual field increase as a function of
eccentricity from the fixation location. As the pooling regions get larger
towards the periphery, more information is lost at these
locations, which  might seem to be a disadvantage, however, the exploration of
the scene with the high resolution fovea through a guided search algorithm might
mitigate the apparent loss of peripheral information. On the other hand, fewer
computational resources
are allocated to process these low resolution areas which, in turn, lower the
computational cost.  In this paper, we investigate the
impact of using a foveated visual field on the detection performance and
its computational cost savings. 

\begin{figure}
    \centering
    \includegraphics[scale=.32]{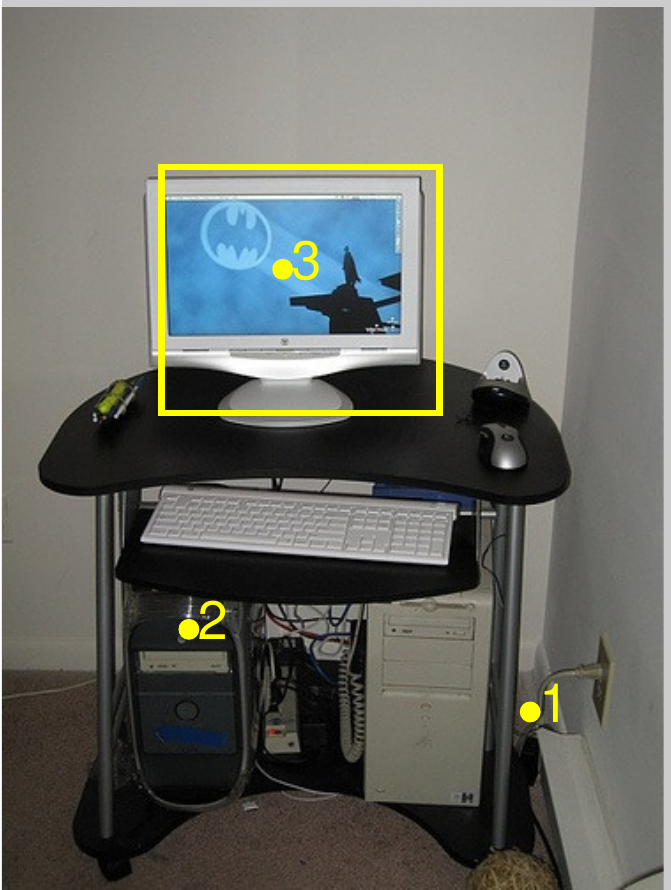} 
    \includegraphics[scale=.33]{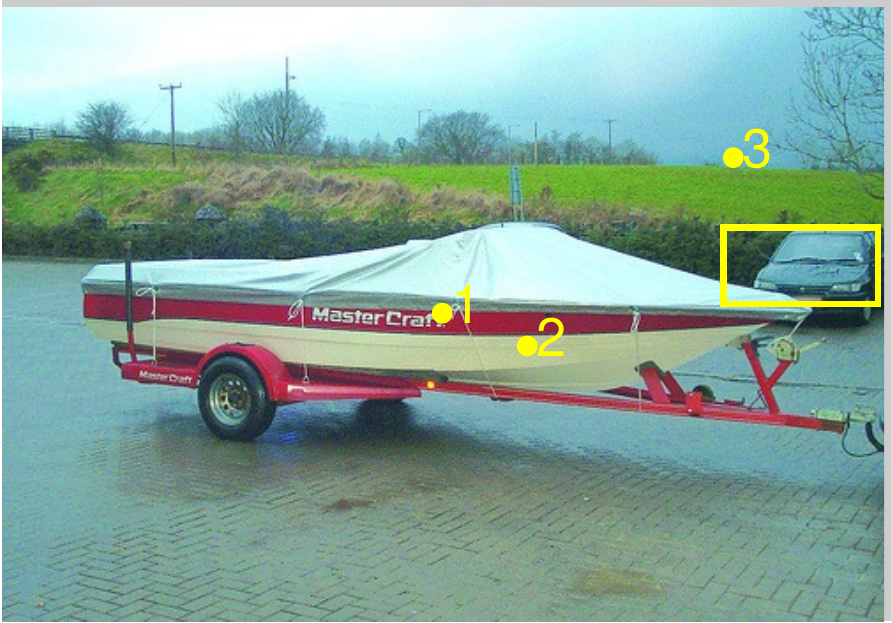} 
    \caption{
        \label{fig:teaser_detection_examples}
    Two example detections by our foveated object
    detector (FOD). Yellow dots  show fixation points, numbers in yellow fonts
    indicate the sequence of fixations and the bounding box is  the final
    detection. Note that FOD does not have to fixate on the target object in
    order to localize it (example on the right).
    }
\end{figure}

\subsection{Overview of our approach}
The foveated object detector (FOD) mimics the process by which humans  search
for objects in scenes utilizing eye movements to point the high resolution fovea
to points of interest (Figure \ref{fig:teaser_detection_examples}). The FOD gets
assigned an initial fixation point on
the input image and collects information by extracting image features through
its foveated visual field. The features extracted around the fixation point are
at fine spatial scale while features extracted away from the fixation location
at coarser scale. This fine-to-coarse transition is dictated by the pooling
region sizes of the visual field. Then, based on the information collected, the
FOD chooses the next fixation point and makes a saccade to that point. Finally,
the FOD integrates information collected through multiple saccades and outputs
object detection predictions.

Training such an object detector entails learning templates at all
locations in the visual field. Because the visual field has varying resolution,
the appearance of a target object varies depending on where it is located within
the visual field. We use the HOG \cite{dalal_triggs:cvpr2005} as image features
and a simplified version of the V1 model \cite{freeman_simoncelli:metamers} to
compute pooled features within the visual field. A mixture of linear
templates is trained at selected locations in the visual field using a
latent-SVM-like \cite{dpm:pami2010, bharath:eccv2012} framework.

\subsection{Contribution} We present an object detector that has a foveated
visual field based on physiological measurements in primate visual cortex
\cite{freeman_simoncelli:metamers} and that models the appearance of target
objects not only in the high resolution fovea but also in the periphery. .
Importantly, the model is developed in the context of a modern
object detection algorithm and a standard data-set (PASCAL VOC) allowing
for the first time direct evaluation of the impact of a foveated visual system
on an object detector.

We believe that object detection using a foveated visual field offers a novel
and promising direction of research in the quest for an efficient alternative to
the sliding window method, and also a possible explanation for why foveated
visual systems might have evolved in organisms. We show that our method achieves
greater computational savings than a state-of-the-art cascaded detection method. 
Another contribution of our work
is the latent-LDA formulation (Section \ref{sec:latent-LDA}) where linear
discriminant analysis is used within a latent-variable learning framework.

In the next section, we describe the FOD in detail and report experimental
results in Section \ref{sec:experiments} which is followed by the related work
section, conclusions and discussion. 

% #########################################################################
\section{The Foveated Object Detector (FOD)}

\subsection{Foveated visual field} 
The Freeman-Simoncelli (FS) model \cite{freeman_simoncelli:metamers} is neuronal
population model of V1 and V2 layers of the visual cortex. The model specifies
how responses are pooled (averaged together) hierarchically beginning from the
lateral geniculate nucleus to V1 and then the V2 layer. V1 cells encode information about
local orientation and spatial frequency whereas the cells in V2 pools V1
responses non-linearly to achieve selectivity for compound features such as
corners and junctions. The model is based on findings and physiological
measurements of the primate visual cortex and specifies the shapes and sizes of
the receptive fields of the cells in V1 and V2.
According to the model, the
sizes of receptive fields increase linearly as a function of the distance from
the fovea and this rate of increase in V2 is larger than that of V1, which means
V2 pools larger areas of the visual field in the periphery. The reader is
referred to \cite{freeman_simoncelli:metamers} for further details.

We simplify the FS model in two ways. First, the model uses a Gabor filter bank
to compute image features and we replace these with the HOG features
\cite{dalal_triggs:cvpr2005,dpm:pami2010}. Second, we
only use the V1 layer and leave the non-linear pooling at V2 as future work. We
use this simplified FS model as the foveated visual field of our object
detector which is shown in  Figure \ref{fig:teaser_foveated_model}. The fovea
subtends a radius of $2$ degrees. We also only simulate a visual field with a
radius of $10$ degrees which is sufficient to cover the test images presented at
a typical viewing distance of $40$ cm. The square
boxes with white borders (Figure \ref{fig:teaser_foveated_model} represent the
pooling regions within the fovea. The
surrounding colored regions are the peripheral pooling regions. While the foveal
regions have equal sizes, the peripheral regions grow in size as a function
-- which is specified by the FS model --  of
their distance to the center of the fovea. The color represents the weights that
are used in pooling, i.e. weighted summation of, the underlying responses. A
pooling region
partly overlaps  with its neighboring pooling regions (see the supplementary
material of \cite{freeman_simoncelli:metamers} for details).  Assuming a viewing
distance of $40$cm, the whole visual field covers about a $500$x$500$ pixel
area (a pixel subtends $0.08^\circ$). The foveal radius is $52$ pixels subtending a visual
angle of $4$ degrees. 

Given an image and a fixation point, we first compute the gradient at each pixel
and then for each pooling region, the gradient magnitudes are pooled per
orientation for the pixels that fall under the region. 
At the  fovea, where the pooling regions are
$8$x$8$ pixels, we use the HOG features at the same spatial scale of the
original DPM model\cite{dpm:pami2010}, and in the periphery, each
pooling region takes a weighted sum of HOG features of the $8$x$8$ regions
that are covered by that  pooling region.

\begin{figure*}
    \centering
    \subfigure[]{
\includegraphics[scale=.22]{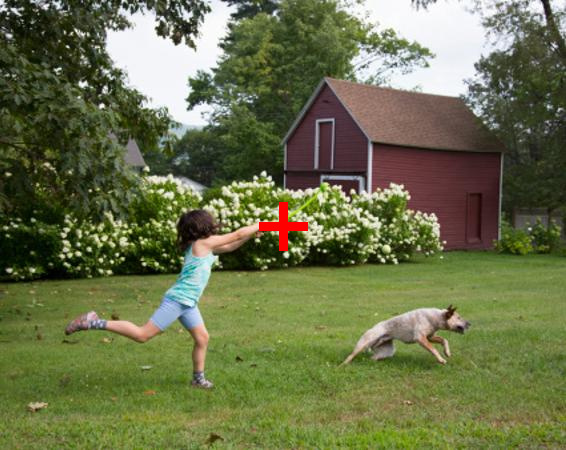}
    }
    \subfigure[]{
\includegraphics[scale=.24]{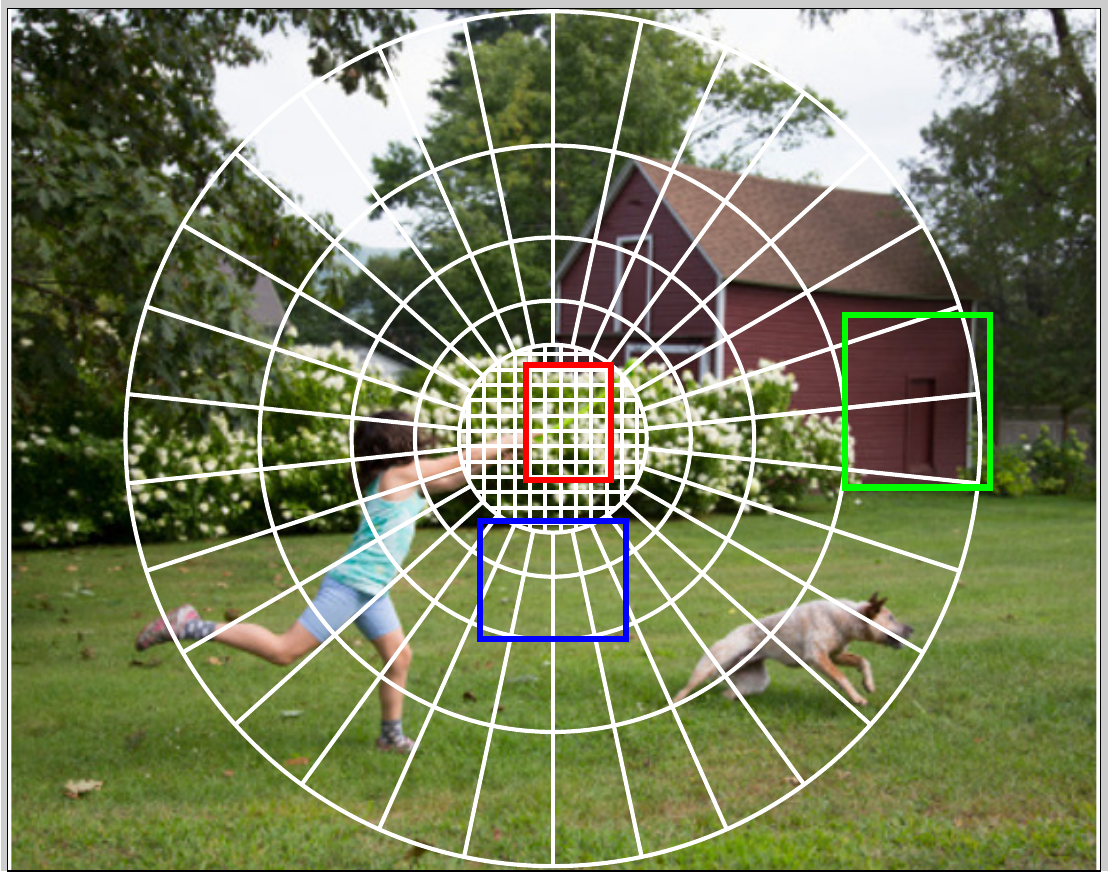}
    }
    \subfigure[]{
\includegraphics[scale=.1674]{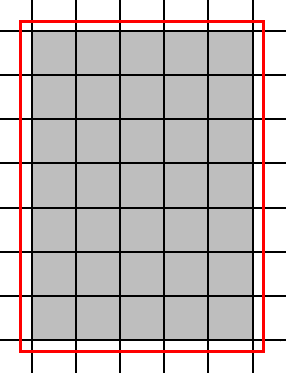}
    }
    \subfigure[]{
\includegraphics[scale=.1247]{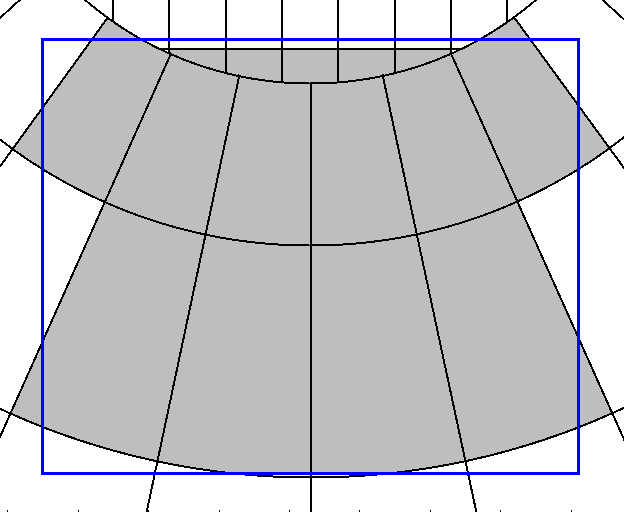}
    }
    \subfigure[]{
\includegraphics[scale=.184]{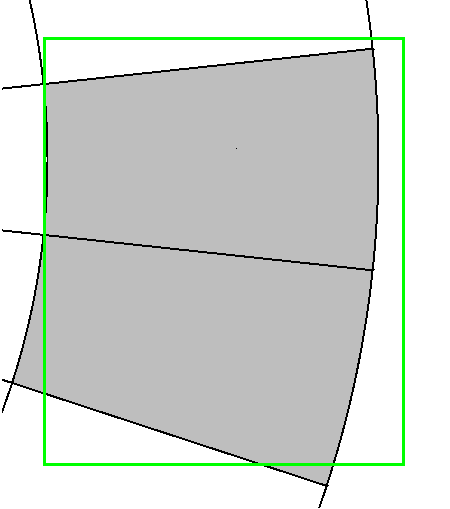}
    }
\caption{
    Illustration of the visual field of the model. (a) The model is fixating at
    the red cross mark on the image. (b) Visual field (Figure
    \ref{fig:teaser_foveated_model}) overlaid on the image, centered at the
    fixation location. White line delineate the borders of pooling
    regions. Nearby pooling regions do overlap. The weights (Figure
    \ref{fig:teaser_foveated_model}) of a pooling region sharply decrease outside
    of its shown borders. White borders are actually iso-weight contours for
    neighboring regions. Colored bounding boxes show the templates of three
    components on the visual field: red, a template within the fovea;  blue and
    green, two peripheral templates at 2.8 and 7 degree periphery, respectively.
    (c,d,e) Zoomed in versions of the red (foveal), blue (peripheral) and green
    (peripheral) templates. The weights of a template, ${\bf w}_i$, are defined
    on the gray shaded pooling regions.
    \label{fig:model_components}
}
\end{figure*}

\subsection{The model}
The model $M$ consists of a mixture of $n$ components
\begin{equation}
    M = \{({\bf w}_i, {\bf \Bell}_i) : i =1,2,\dots,n\}
    \label{eq:model_components}
\end{equation}
\noindent where ${\bf w}_i$ is a linear template and
$\Bell_i$ is the location of the template with respect to the center of the
visual field. The location variable $\Bell_i$ defines a unique bounding box
within the visual field for the $i^\mathrm{th}$ template. Specifically,  
$\Bell_i = (\omega_i, h_i, x_i, y_i)$  is a $4$-tuple
whose variables respectively denote width, height and $x$,$y$ coordinates of the
$i^\mathrm{th}$ template within the visual field.
The template, ${\bf w}_i$, is a matrix of weights on the features extracted from
the
pooling regions underlying the bounding box $\Bell_i$. The dimensionality of
${\bf w}_i$, i.e.  the total number of weights, depends both on the width and
height of its bounding box and its location in the visual field. A component
within the fovea covers a larger number of pooling regions compared to a
peripheral component
with the same width and height, hence the dimensionality of a foveal template is
larger. Three example components are illustrated in Figure
\ref{fig:model_components} where the foveal
component (red) covers $7$x$5=35$ pooling regions while the (blue and green)
peripheral components cover $15$ and $2$ regions, respectively.  Since a fixed
number of features\footnote{We use the feature extraction implementation of DPM
(rel5) \cite{dpm-voc-release5, dpm:pami2010}, which extracts a $31$-dimensional
feature vector.}  is extracted
from each pooling region (regardless of its
size), foveal components have higher-resolution templates associated with them.

\subsubsection{Detection model}
Suppose that we are given a model $M$ that is already trained for a certain
object class. The model is presented with an image $I$ and assigned an initial
fixation location $\bs f$.  We are interested in searching for an object instance in
$I$. Because
the size of a searched object is not known apriori, the model has to analyze the
input image at various scales. We use the same set of image scales given in
\cite{dpm:pami2010} and use $\sigma$ to denote a scale from that set. When used
as a subscript to
an image, e.g. $I_\sigma$, it denotes the scaled version of that image, i.e.
width (and height) of $I_\sigma$  is $\sigma$  times the width (and height) of
$I$. $\sigma$ also applies to
fixation locations and bounding boxes: if $\bs f$ denotes a fixation location
$(\bs f_x, \bs f_y)$, then $\bs f_\sigma = (\sigma \bs f_x, \sigma \bs f_y)$;
for a bounding box $\bs b=(w,h,x,y)$,  $\bs b_\sigma=(\sigma w,\sigma h,\sigma
x,\sigma y)$. 

To check whether an arbitrary bounding box $\bs b$ within $I$ contains an object
instance, while the model is fixating at location f, we compute a detection
score as

\begin{equation}
    \mathrm{s}(I,\bs b, \bs f) = \underset{\sigma}{\mathrm{max}} \underset{c
    \in G(\bs b_\sigma, \bs f_\sigma)}{\mathrm{max}}\;\rmbf{w}^T\Psi(I_\sigma,\bs
    f_\sigma, c)
\end{equation}

\noindent where $\Psi(I_\sigma,\bs f_\sigma, c)$ is a feature extraction
function which returns the features of $I_\sigma$ for component $c$ (see
Equation \eqref{eq:model_components}) when the
model is fixating at $\bs f_\sigma$. The vector $\rmbf w$ is the blockwise
concatenation of the templates of all components. $\Psi(\cdot)$  effectively
chooses which component to use, that is $ \rmbf{w}^T\Psi(I_\sigma,\bs
    f_\sigma, c)=\rmbf{w}_c^T\Psi(I_\sigma,\bs
    f_\sigma, c)$ .  The fixation location ,$\bs f_\sigma$,  together
with the component $c$ define a unique location, i.e. a bounding box, on
$I_\sigma$.  $G(\bs b_\sigma,\bs f_\sigma)$ returns the set of all components
whose templates have a
predetermined overlap (intersection over union should be at least $0.7$ as 
in \cite{dpm:pami2010}) with $\bs b_\sigma$
when the model
is fixating at $\bs f_\sigma$. During both training and testing, $\sigma$ and
$c$ are latent variables for example $(I,\bs b)$.

Ideally, $s(I,\bs b,\bs f) > 0$ should hold for an appropriate $\bs f$ when $I$
contains an object instance within $\bs b$. For an image that does not contain
an object instance, $s(I,\bs b = \emptyset, \bs f) < 0$ should hold for any $\bs
f$. For this to work, a subtlety in $G(\cdot)$'s definition is
needed: $G(\emptyset,\bs f)$ returns all components of the model (Equation
\eqref{eq:model_components}). During
training (Section \ref{sec:model_training}), this will enforce the responses of all components
for a negative image to be suppressed down.

\subsubsection{Integrating observations across multiple fixations}
So far, we have looked at the situation where the model has made only one
fixation. We describe in Section \ref{sec:map_model} how the model chooses the
next fixation location. For now, suppose that the model has made $m$ fixations,
$\bs f_1,\bs f_2,\dots,\bs f_m$, and we want to find out whether an arbitrary
bounding box $\bs b$ contains an object instance. This computation involves
integrating observations across multiple fixations, which is a considerably more
complicated problem than the single fixation case. The Bayesian decision on
whether $\bs b$ contains an object instance is based on the comparison of
posterior probabilities:

\begin{equation}
    \label{eq:posterior_comparison}
    \frac{
    P(y_{\bs b}=1|\bs f_1,\bs f_2,  \dots, \bs f_m,I) }
    {P(y_{\bs b}=0|\bs f_1,\bs f_2,  \dots, \bs f_m,I) }
    \;\substack{< \\ >} \;1
\end{equation}

where $y_{\bs b} = 1$ denotes the event that there is an object instance at
location $\bs b$.
We use the posteriors' ratio as a detection score, the higher it is the more
likely $\bs b$ contains an instance. Computing the probabilities in
\eqref{eq:posterior_comparison} requires
training a classifier per combination of fixation locations for each different
value of $m$, which is intractable. We approximate it using a conditional
independence assumption (derivation given in Appendix \ref{apx:bayesian_decision}):

\begin{equation}
\label{eq:FOD_decision}
    \frac{
    P(y_{\bs b}=1|\bs f_1,\bs f_2,  \dots, \bs f_m,I) }
    {P(y_{\bs b}=0|\bs f_1,\bs f_2,  \dots, \bs f_m,I) } \approx 
    \prod_{i=1}^m \frac{P(y_{\bs b}=1 | \bs f_i, I)}
    {P(y_{\bs b}=0 | \bs f_i, I)}.
\end{equation}

We model the probability $P(y_{\bs b}=1|\bs f,I)$ using a classifier and use the
sigmoid transfer function to convert raw classification scores to probabilities:

\begin{equation}
    \label{eq:score_to_posterior}
    P(y_{\bs b}=1|\bs f,I) = \frac{1}{1+e^{(-s(I,\bs b, \bs f))}}.
\end{equation}

We simplify the computation in \eqref{eq:FOD_decision} by taking the log
(derivation given in Appendix \ref{apx:sum_of_scores}):
\begin{equation}
    \label{eq:sum_of_scores}
    \mathrm{log}\left(\prod_{i=1}^m \frac{P(y_{\bs b}=1 | f_i, I)}
    {P(y_{\bs b}=0 | f_i, I)} \right) = \sum_{i=1}^m s(I,\bs b, \bs f_i).
\end{equation}

Taking the logarithm of posterior ratios does not alter the ranking of detection
scores for different locations, i.e. $\bs b$'s, because logarithm is a monotonic
function. In short, the detection score computed by the FOD for a certain
location $\bs b$, is the
sum of the individual scores for $\bs b$ computed at each fixation. 

After
evaluating \eqref{eq:sum_of_scores} for a set of candidate locations, final
bounding box predictions are obtained by non-maxima suppression
\cite{dpm:pami2010}, i.e. given multiple predictions for a certain location,
all predictions except the one with the maximal score are discarded.

\subsection{Eye movement strategy}
\label{sec:map_model}
We use the maximum-a-posteriori (MAP) model \cite{beutter:2003} as the basic eye
movement strategy of the FOD. The MAP
model is shown to be consistent with human eye movements in a variety of visual
search tasks \cite{beutter:2003, verghese:visionres2012}. Studies have
demonstrated that in some circumstances human saccade statistics better match an
ideal searcher \cite{najemnik:nature2005} that makes eye movements to locations
that maximize the accuracy of localizing targets, yet in many circumstances the
MAP model approximates the ideal searcher \cite{Najemnik2009, zhang:2010}
but is computationally more tractable for objects in real scenes.  The MAP model
select the location with the highest posterior probability of containing the
target object as the next fixation location, that is $\bs f_{i+1} =$ center of
$\Bell^*$ where 

\begin{equation}
\Bell^*=\underset{\Bell}{\operatorname{arg}\,\operatorname{max}}\;
P(y_{\Bell}=1 | \bs f_1, \bs f_2,\dots,\bs f_i, I).
\end{equation}

\noindent Finding the maximum of the posterior above is equivalent to finding
the maximum of the posterior ratios,

\begin{equation}
    \textstyle
\underset{\Bell}{\operatorname{arg}\,\operatorname{max}}\;
P(y_{\Bell}=1 | \bs f_1, \dots,\bs f_i, I) = 
\underset{\Bell}{\operatorname{arg}\,\operatorname{max}}\;
\frac{P(y_{\Bell}=1 | \bs f_1, \dots,\bs f_i, I)}
{P(y_{\Bell}=0 | \bs f_1, \dots,\bs f_i, I)}
\end{equation}

\noindent since for two arbitrary locations $\Bell_1,\Bell_2$; let
$p_1=P(y_{\Bell_1}=1|\cdot)$ and $p_2=P(y_{\Bell_2}=1|\cdot)$, then we have

\begin{equation}
    \frac{p_1}{1-p_1}> \frac{p_2}{1-p_2} \implies p_1 > p_2.% \; \; \; \; \forall p_1, p_2 \in [0,1].
\end{equation}

%\begin{wrapfigure}[]{l}{0.17\textwidth}
\begin{figure}
    \centering
\includegraphics[scale=.3]{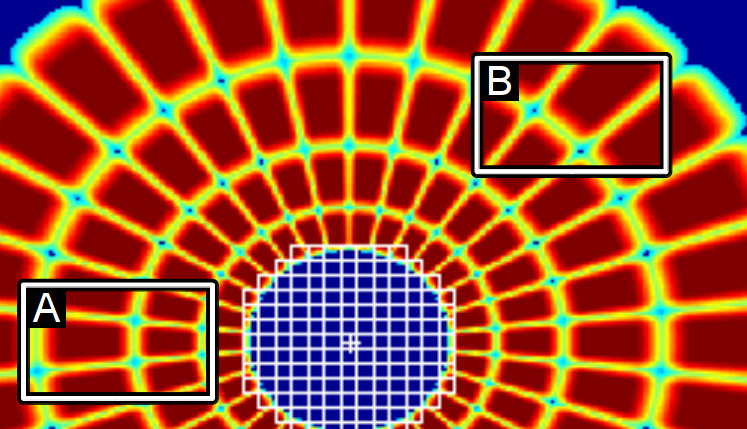}
\caption{
Two bounding boxes (A,B) are shown on the visual field. While box A covers a large
portion of the pooling regions that it intersects with, box B's coverage is
not as good. Box B is discarded
as it does not meet the overlap criteria (see text), therefore a component for B
in the model is not created. 
\label{fig:good_boxes_bad_boxes}
}
\end{figure}
%\end{wrapfigure}

\subsection{Training the model}
\label{sec:model_training}
\subsubsection{Initialization}
\label{sec:model_initialization}
A set of dimensions (width and height) is determined from the bounding box
statistics of the examples in the training set as done in the initialization of
the DPM model \cite{dpm:pami2010}. Then, for each width and height, new
components
with these dimensions are created to tile the entire visual field. However, the
density of components in the visual field is not uniform. Locations, i.e.
bounding boxes, that do not overlap well with the underlying pooling regions
are discarded. To define goodness of overlap, a bounding box is said to
intersect with an underlying pooling region if more than one fifth of that
region is covered by the bounding box. Overlap is the average coverage across
the intersected regions. If the overlap is more than $75\%$, then a component for
that location is created, otherwise the location is discarded (see Figure
\ref{fig:good_boxes_bad_boxes} for
an example). In addition, no components are created for locations that are
outside of the visual field. Weights of the component templates (${\bf w}_i$) are
initialized to arbitrary values. Training the model is essentially optimizing
these weights on a given dataset.

\subsubsection{Training}
\label{sec:latent-LDA}
Consider a training set $\mathcal{D} = \{ (I_i,\bs b_i) \}_{i=1}^K$
where $I_i$ is an image and $\bs b_i$ a bounding box and $K$ is the total number
of examples. 
If $I_i$
does not contain any positive examples, i.e. object instances, then $\bs
b_i=\emptyset$. Following the DPM model \cite{dpm:pami2010}, we train model
templates using a latent-SVM formulation: 

\begin{equation}
    \label{eq:latent_svm_cost}
    \underset{\rmbf w}{\arg\min}\; \frac{1}{2}||\rmbf w||_2^2 + C\sum_{i=1}^K
    \sum_{\bs f \in F(I_i, \bs b_i)}
    \mathrm{max}(0, 1-y_i s(I_i, \bs b_i, \bs f)).
\end{equation}
 
\noindent where $y_i=1$ if $\bs b_i \neq \emptyset$ and $y_i=-1$, otherwise.
The set $F(I_i, \bs b_i)$ denotes the set of all \emph{feasible} fixation
locations for example $(I_i, \bs b_i)$. For $\bs b_i \neq \emptyset$, a fixation
location is considered feasible if there exists
a model component whose bounding box overlaps with $\bs b_i$. For $\bs
b_i=\emptyset$, all possible fixation locations on $I_i$ are considered
feasible.

Optimizing the cost function in \eqref{eq:latent_svm_cost} is
manageable for mixtures with few components, however, the FOD has a large number
of components in its visual field (typically, for an object class in the PASCAL VOC
2007 dataset \cite{pascal-voc-2007}, there are around $500$-$700$)
and optimizing this cost function becomes prohibitive in terms of computational
cost. As an alternative,
cheaper linear classifiers can be used. Recently, linear discriminant analysis
(LDA) has been used in object detection (\cite{bharath:eccv2012}) producing
surprisingly good results with much faster training time.
Training a LDA classifier amounts to computing $\Sigma^{-1}(\mu_1 - \mu_0)$
where $\mu_1$ is the mean of the feature vectors of the positive
examples, $\mu_0$ is the same for the negative examples and  $\Sigma$ is the
covariance matrix of these features. Here, the most expensive computation is the
estimation of $\Sigma$, which is required for each template with different
dimensions.
However, it is possible to estimate a
global $\Sigma$ from which covariance matrices for templates of different
dimensions can
be obtained \cite{bharath:eccv2012}. For the FOD, we estimate the covariance
matrices for the foveal templates and estimate the 
covariance matrices for peripheral templates by applying the feature pooling
transformations to the foveal covariance matrices. 

We propose to use LDA in a latent-SVM-like framework as an alternative to the
method in \cite{bharath:eccv2012} where positive examples are clustered first
and then a LDA
classifier is trained per cluster. Consider the $t^\mathrm{th}$ template, ${\bf
w}_t$. LDA gives us
that  LDA gives us that $\rmbf w_{t,\text{LDA}} = \Sigma_t^{-1}
(\mu_t^{\mathrm{pos}} - \mu_t^{\mathrm{neg}})$ where $\Sigma_t$ is the
covariance matrix for template $t$, $\mu_t^{\mathrm{pos}}$ and
$\mu_t^{\mathrm{neg}}$ are the mean of positive and negative feature vectors,
respectively, assigned to template $t$. We propose to apply an affine
transformation to the LDA classifier:

\begin{equation}
    \rmbf w_t = 
    \left[ \begin{smallmatrix}
          \alpha_t\\
              & \alpha_t & & \text{\large0}\\
              & & \ddots\\
              & \text{\large0} & & \alpha_t\\
              & & & & \beta_t
  \end{smallmatrix} \right]
     \begin{bmatrix}
\rmbf w_{t,\text{LDA}} \\ 1
     \end{bmatrix} = 
     \begin{bmatrix}
         \alpha_t \rmbf w_{t,\text{LDA}} \\ \beta_t
     \end{bmatrix}
\end{equation}

\noindent and modify the cost function as

\begin{equation}
    \label{eq:latent-LDA}
    \begin{array}{c}
        \\
\underset{\bs\alpha,\bs\beta}{\arg\min}\bigg(
    \frac{1}{2}||\rmbf w||_2^2 +
C\sum_{t=1}^{N}  \mathrm{max}(0,1+\rmbf{w_t}^T \mu^{neg}_t) + \\
C\sum_{i\in\{i|\bs b_i \neq\emptyset\}} \sum_{\bs f \in F(I_i, \bs b_i)}
    \mathrm{max}(0, 1-y_i s(I_i, \bs b_i, \bs f)) 
\bigg)
\end{array}
\end{equation}

\noindent where the first summation pushes the score of the mean of the negative
examples to under zero and the second summation, taken over positive examples
only, pushes the scores to above 0. $\bs\alpha$ and $\bs\beta$ are appropriate
blockwise concatenation of $\alpha_t$ and $\beta_t$s. $C$ is the regularization
constant. Overall,
this optimization effectively calibrates the dynamic ranges of different
templates' responses  in the model so that the scores of positive examples and negative
means are pushed away from each other while the norm of $\rmbf w$ is constraint
to prevent overfitting. This formulation does not require the costly
mining of hard-negative examples of latent-SVM.  
We call this formulation (Equation \eqref{eq:latent-LDA}) as latent-LDA.

To optimize \eqref{eq:latent-LDA}, we use the classical coordinate-descent
procedure. We start by initializing $\rmbf w$ by training on warped-positive
examples as in \cite{dpm:pami2010}. Then, we alternate between  choosing the best
values for the latent variables  while keeping $\rmbf w$ fixed, and optimizing
for $\rmbf w$ while keeping the latent variables of positive examples fixed.

% #########################################################################
\section{Experiments}
\label{sec:experiments}
We evaluated our method on the PASCAL VOC 2007 detection ({\tt comp3})
challenge dataset and protocol (see \cite{pascal-voc-2007} for details). All
results are obtained by training on the {\tt train+val} split and testing on the
{\tt test} split.

\begin{table*}
    \caption{Average precision (AP) scores of SW based methods on the PASCAL
    VOC 2007 dataset.}
    \label{table:SW_ap_results}
\centering
\begin{tabular}{c|c|*{20}{c}|c}
\multicolumn{2}{c|}{} &\bf aero& \bf bike& \bf bird& \bf boat& \bf bottle& \bf bus& \bf car& \bf cat& \bf chair& \bf cow& \bf table& \bf dog& \bf horse& \bf mbike& \bf person& \bf plant& \bf sheep& \bf sofa& \bf train& \bf tv& \bf mAP \\
\hline
\multicolumn{2}{c|}{\bf DPM \cite{dpm:pami2010}} &
23.6 & 48.6 & 9.7 & 11.0 & 19.3 & 40.4 & 45.2 & 12.4 & 15.4 & 19.4 & 17.4 & 4.0 & 44.7 & 36.4 & 31.2 & 10.9 & 14.1 & 19.5 & 32.2 & 37.0 &  24.6 \\
\hline
\multicolumn{2}{c|}{\bf E-SVM \cite{malisiewicz:iccv2011}} &
20.4 & 40.7 & 9.3 & 10.0 & 10.3 & 31.0 & 40.1 & 9.6 & 10.4 & 14.7 & 2.3 & 9.7 & 38.4 & 32.0 & 19.2 & 9.6 & 16.7 & 11.0 & 29.1 & 31.5 &  19.8 \\
\hline
\multicolumn{2}{c|}{\bf DCC \cite{bharath:eccv2012}} &
17.4 & 35.5 & 9.7 & 10.9 & 15.4 & 17.2 & 40.3 & 10.6 & 10.3 & 14.3 & 4.1 & 1.8 & 39.7 & 26.0 & 23.1 & 4.9 & 14.1 & 8.7 & 22.1 & 15.2 &  17.1 \\
\hline
\multicolumn{2}{c|}{\bf Our SW} &
17.5 & 28.6 & 9.7 & 10.4 & 17.3 & 29.8 & 36.7 & 7.9 & 11.2 & 21.0 & 2.3 & 2.7 & 30.9 & 21.1 & 19.7 & 3.0 & 9.2 & 13.7 & 23.5 & 25.2 &  17.1 \\
\hline
\end{tabular}
\end{table*}

\subsection{Comparison of SW based methods} 
We first compared our SW implementation, which corresponds to using foveal
templates only, to three state-of-the-art methods that
are also SW based \cite{dpm:pami2010, malisiewicz:iccv2011, bharath:eccv2012}.
Table \ref{table:SW_ap_results} gives
the  AP (average precision) results, i.e. area under the precision-recall curve
per class, and mean AP (mAP) over all classes.
Originally, the deformable parts model (DPM)  uses object parts, however, in
order to make  a fair comparison with our model, we disabled its parts. The
first row of Table \ref{table:SW_ap_results} shows the latest version of the  DPM system
\cite{dpm-voc-release5} with  the parts-learning code disabled. The second row
shows results for another popular SVM-based system, known as the exemplar-SVM
(E-SVM), which also only models whole objects, not its parts.  Finally, the
third row shows results from a LDA-based system, ``discriminative decorrelation
for classification" (DCC) \cite{bharath:eccv2012}. All three systems are based
on HOG features and mixture of linear templates.  The results show that SVM
based systems perform
better than the LDA based systems, which is not a surprising finding since it is
well known that discriminative models outperform generative models in
classification tasks. However, LDA's advantage against this performance loss is
that it is ultra fast to train, which is exactly the reason we chose to use LDA
instead of SVM. Once the background covariance matrices are estimated (which can
be done once and for all \cite{bharath:eccv2012}), training is as easy as taking
the average of the feature vectors of positive examples and doing a matrix
multiplication.  We estimated the time that training a SVM based system for our
FOD to be about 300 hours (approximately 2 weeks) for a single object class, whereas
the LDA based system can be trained under an hour on the same machine which has
an Intel i7 processor. 

Although our SW method achieves the same mean AP (mAP) score  as the DCC method
\cite{bharath:eccv2012}, the latter has a detection model with higher
computational cost. We use 2 templates per class  while DCC trains
more than 15 templates per class within an
exemplar-SVM\cite{malisiewicz:iccv2011}-like framework. DCC considers the dot
product of the feature vector
of the detection window with every exemplar within a cluster,
which basically means that a detection window is compared to all positive
examples in the training set. In our case, the number of dot products considered
per detection window is equal to the number of templates, which is 2 in this
paper, which clearly demonstrates the advantage of our latent-LDA approach over
DCC \cite{bharath:eccv2012}.
%There are other variants of DCC in \cite{bharath:eccv2012} with more
%complicated detection models, e.g. with context, 

\subsection{Comparison of FOD with SW}
Next, we compared the performance of FOD with our SW method. We
experimented with two eye movement strategies, MAP (Section \ref{sec:map_model}) and
random strategy to demonstrate the importance of guidance of eye movements.

\begin{table*}
    \caption{AP scores and relative computational costs of SW and FOD on the PASCAL VOC 2007 dataset.}
    \label{table:FOD_ap_results}
\centering
%\scriptsize
\begin{tabular}{c|c|*{20}{c}|c||c}
%\multicolumn{2}{c|}{} &\rotatebox[origin=c]{50}{\bf aero}& \rotatebox[origin=c]{50}{\bf bike}& \rotatebox[origin=c]{50}{\bf bird}& \rotatebox[origin=c]{50}{\bf boat}& \rotatebox[origin=c]{50}{\bf bottle}& \rotatebox[origin=c]{50}{\bf bus}& \rotatebox[origin=c]{50}{\bf car}& \rotatebox[origin=c]{50}{\bf cat}& \rotatebox[origin=c]{50}{\bf chair}& \rotatebox[origin=c]{50}{\bf cow}& \rotatebox[origin=c]{50}{\bf table}& \rotatebox[origin=c]{50}{\bf dog}& \rotatebox[origin=c]{50}{\bf horse}& \rotatebox[origin=c]{50}{\bf mbike}& \rotatebox[origin=c]{50}{\bf person}& \rotatebox[origin=c]{50}{\bf plant}& \rotatebox[origin=c]{50}{\bf sheep}& \rotatebox[origin=c]{50}{\bf sofa}& \rotatebox[origin=c]{50}{\bf train}& \rotatebox[origin=c]{50}{\bf tv}& \rotatebox[origin=c]{50}{\bf mAP} \\
\multicolumn{2}{c|}{} &\bf aero& \bf bike& \bf bird& \bf boat& \bf bottle& \bf
                   bus& \bf car& \bf cat& \bf chair& \bf cow& \bf table& \bf
                   dog& \bf horse& \bf mbike& \bf person& \bf plant& \bf sheep&
              \bf sofa& \bf train& \bf tv& \bf mAP & \bf Comp. Cost\\
\hline
\multicolumn{2}{c|}{\bf Our SW} &
17.5 & 28.6 & 9.7 & 10.4 & 17.3 & 29.8 & 36.7 & 7.9 & 11.2 & 21.0 & 2.3 & 2.7 &
30.9 & 21.1 & 19.7 & 3.0 & 9.2 & 13.7 & 23.5 & 25.2 &  17.1 & 100 \\
\hline
\hline
\multirow{3}{*}{\bf MAP-C} &\bf 1 &
17.0 & 21.1 & 4.9 & 9.8 & 9.3 & 27.4 & 27.9 & 8.5 & 3.7 & 12.8 & 2.0 & 4.3 &
29.7 & 19.7 & 18.2 & 1.2 & 10.7 & 14.0 & 26.2 & 21.8 &  14.5 &  11.5\\
 & \bf 3 &
17.4 & 27.7 & 10.1 & 10.6 & 10.4 & 30.8 & 31.6 & 8.4 & 10.4 & 17.2 & 2.1 & 3.4 &
33.3 & 21.1 & 18.7 & 3.4 & 7.6 & 15.4 & 26.4 & 23.5 &  16.5 & 31.2 \\
 & \bf 5 &
17.0 & 28.6 & 10.0 & 10.7 & 11.2 & 31.0 & 34.0 & 8.3 & 10.6 & 18.2 & 2.1 & 3.4 &
34.2 & 21.8 & 19.7 & 2.8 & 8.1 & 15.1 & 27.8 & 24.0 &  16.9 & 49.6 \\
\hline
\multirow{3}{*}{\bf MAP-E} &\bf 1 &
1.6 & 7.1 & 4.1 & 5.6 & 9.1 & 8.7 & 11.7 & 6.0 & 3.6 & 10.2 & 2.0 & 2.2 & 8.5 &
  10.2 & 13.5 & 1.3 & 6.8 & 8.0 & 10.6 & 10.3 &  7.1 & 8.7  \\
 & \bf 3 &
13.0 & 24.6 & 9.9 & 9.8 & 10.7 & 27.2 & 29.3 & 7.4 & 10.4 & 16.4 & 3.7 & 2.2 &
30.6 & 20.8 & 16.9 & 3.3 & 11.2 & 13.8 & 23.0 & 24.1 &  15.4 & 28.1 \\
 & \bf 5 &
15.1 & 28.0 & 9.9 & 10.4 & 11.6 & 29.9 & 33.0 & 8.3 & 10.6 & 18.7 & 2.7 & 4.1 &
33.7 & 22.6 & 18.9 & 3.1 & 7.1 & 14.7 & 25.5 & 25.2 &  16.7 & 46.9 \\
\hline
\multirow{3}{*}{\bf RAND} &\bf 1 &
8.2 & 9.3 & 5.5 & 9.3 & 7.8 & 12.2 & 16.2 & 6.1 & 6.8 & 7.5 & 1.6 & 2.5 & 10.6 &
9.1 & 9.9 & 1.9 & 5.0 & 6.7 & 11.2 & 10.0 & \tiny 7.9$\pm$1.4 &
\multirow{3}{*}{\begin{tabular}[x]{@{}c@{}}similar\\to above\end{tabular}} \\
 & \bf 3 &
9.6 & 13.0 & 3.2 & 9.6 & 9.3 & 16.9 & 23.5 & 8.8 & 9.4 & 9.9 & 1.8 & 3.2 & 16.5 & 12.3 & 12.2 & 2.7 & 3.9 & 9.3 & 16.9 & 11.7 & \tiny 10.2$\pm$0.9 \\ 
 & \bf 5 &
10.9 & 15.3 & 3.8 & 9.7 & 9.6 & 20.5 & 26.3 & 9.3 & 9.5 & 10.6 & 1.5 & 3.1 & 20.9 & 13.7 & 13.5 & 2.7 & 3.9 & 12.0 & 18.9 & 12.4 & \tiny 11.4$\pm$1.0 \\
\hline
\multirow{3}{*}{\bf RAND-C} &\bf 1 &
\multicolumn{20}{c}{This row is the same with the ``MAP-C, 1" above.}& & 
\multirow{3}{*}{"}\\
 & \bf 3 &
17.5 & 20.4 & 3.7 & 10.0 & 9.3 & 28.6 & 27.4 & 11.5 & 6.7 & 11.8 & 1.7 & 3.5 & 31.7 & 18.0 & 15.4 & 2.7 & 5.4 & 15.2 & 26.1 & 15.8 & \tiny 14.1$\pm$0.5 \\
 & \bf 5 &
17.6 & 21.4 & 5.2 & 9.9 & 9.7 & 28.1 & 28.6 & 11.4 & 9.6 & 12.1 & 1.6 & 3.5 & 30.0 & 17.9 & 15.3 & 3.7 & 6.7 & 14.4 & 25.4 & 15.9 & \tiny 14.4$\pm$0.7 \\
\hline
\multirow{3}{*}{\bf RAND-E} &\bf 1 &
\multicolumn{20}{c}{This row is the same with the ``MAP-E, 1" above.}& & 
\multirow{3}{*}{"}\\
 & \bf 3 &
9.1 & 13.1 & 2.8 & 9.7 & 9.4 & 17.8 & 22.5 & 9.0 & 6.6 & 10.7 & 2.3 & 3.7 & 14.9 & 12.0 & 14.9 & 1.3 & 3.9 & 2.4 & 13.6 & 14.1 & \tiny 9.7$\pm$0.7 \\
 & \bf 5 &
10.7 & 15.9 & 4.1 & 8.7 & 9.5 & 21.9 & 26.0 & 8.2 & 9.7 & 11.6 & 1.7 & 4.3 & 17.6 & 13.7 & 14.1 & 1.9 & 5.7 & 4.8 & 15.7 & 15.8 & \tiny 11.1$\pm$1.1 \\
\hline

\end{tabular}
\end{table*}

Table \ref{table:FOD_ap_results} shows the AP  scores for FOD with different eye
movement strategies and different number of fixations. We also include in this
table the ``Our SW" result from Table \ref{table:SW_ap_results} for ease of
reference. The MAP and random strategies are denoted with MAP and RAND, 
respectively. Because the model accuracy results will depend on initial point
of fixation, we ran the models with different initial points of fixation.  
The presence of a suffix on a model refers to the
location of the initial fixation:  ``-C" stands for the center of the input
image, i.e. $(0.5, 0.5)$ in normalized image coordinates where the top-left
corner is taken as $(0,0)$ and the bottom-right corner is $(1,1)$; and  ``-E"
for the two locations at the left and right edges of the image, $10\%$ of the image width
away from the image border, that is $(0.1,0.5)$ and $(0.9, 0.5)$.
%one at the left edge, the other on the right, both $10\%$ of the image width
%away from the image border.
MAP-E and RAND-E results are the performance average of
two different runs, one with initial fixation close to the left edge of the
image, the other run close to the right edge of the image. For
the random eye movement, we report the $95\%$ confidence interval for AP over $10$ different runs. We ran all systems for
a total of $5$ fixations. Table
\ref{table:FOD_ap_results} shows 
results for after $1$,$3$ and $5$ fixations. A condition with one fixation is a
model that makes decisions based only on the initial fixation. 

The results show that the FOD using the MAP rule with 5 fixations (MAP-C,5 for short)
performs nearly as good as the SW (a difference of $0.2$ in mean AP). 

%\begin{wrapfigure}{l}{0.5\textwidth}
\begin{figure}
    \centering
    %\subfigure[]{
 %\includegraphics[scale=.23]{figures/pascal_bbox_locations}
        %\label{fig:pascal_bbox_locations}
    %}
    %\subfigure[]{
    \includegraphics[scale=.45]{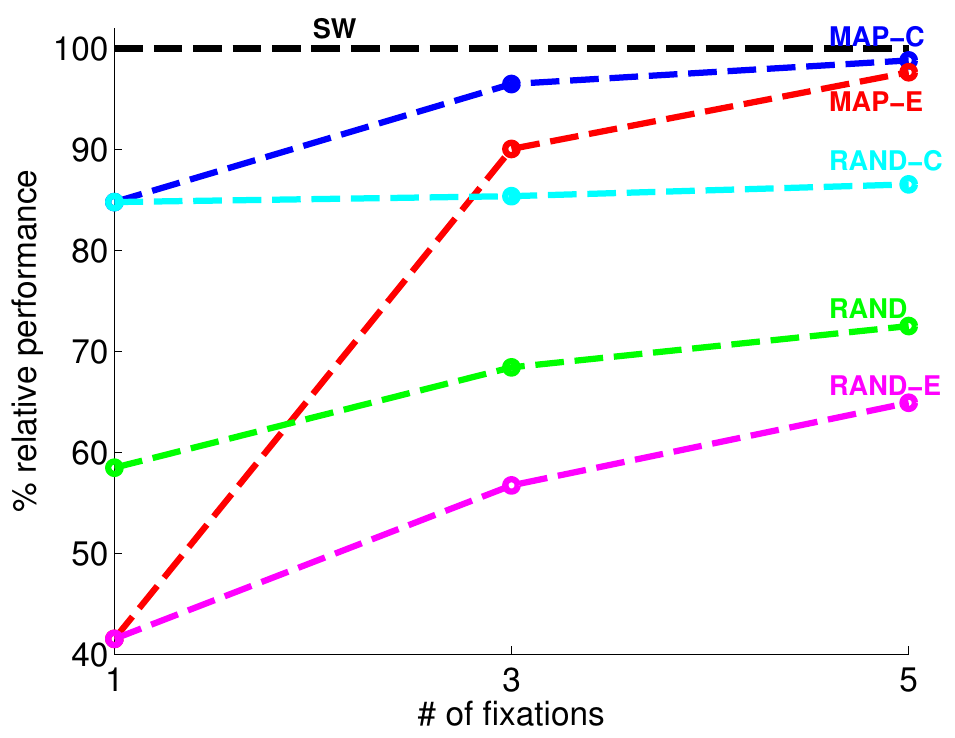} 
    %}
    \caption{
        \label{fig:relative_mAP}
        %(a) Spatial histogram of centers of ground truth bounding in the {\tt
    %train+val} split of the PASCAL VOC 2007 dataset. Centers of bounding boxes
    %concentrate at the image center.
        Ratio of mean AP scores of FOD systems relative to that of  the SW
        system. Graph shows two eye movement algorithms: maximum aposteriori
        probability (MAP) and random (RAND) and two starting points (C:
        center; E: edge). 
    }
%\end{wrapfigure}
\end{figure}

\begin{figure}
    \centering
    \includegraphics[scale=.285]{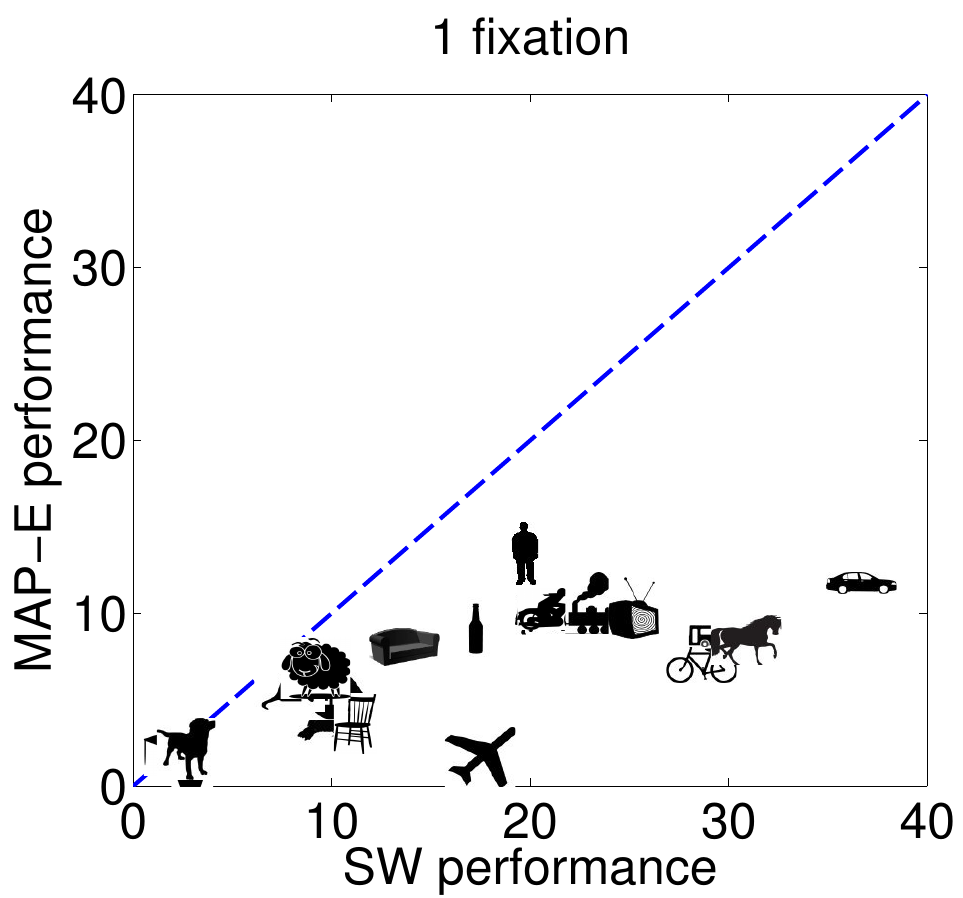} 
    \includegraphics[scale=.285]{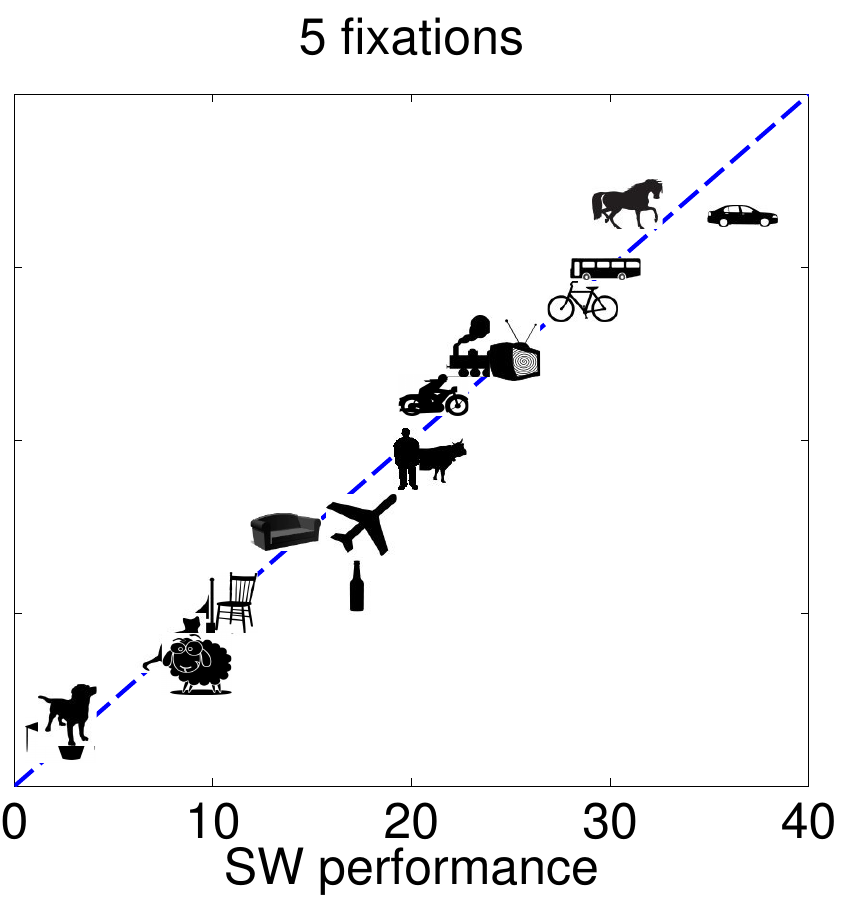}
    \caption{
AP scores achieved by SW and MAP-E per class. 
        \label{fig:MAPE_vs_SW}
    }
%\end{wrapfigure}
\end{figure}

Figure
\ref{fig:relative_mAP} shows the ratio of mean AP for the FOD with the various eye movement
strategies to that of the SW system (relative performance) as a function of
fixation. The relative performance of the MAP-C  to SW (AP of MAP-C divided by
AP of SW) is $98.8\%$ for 5 fixations, $96.5\%$ for 3 fixations and $84.8\%$ for 1
fixation. 
%Figure \ref{fig:MAPE_vs_SW} shows AP scores achieved by SW and FOD, per object
%class. 
The FOD with
eye movement guidance towards the target (MAP-C,5) achieves or exceeds SW's
performance with only 1 fixation in 4 classes, with 3 fixations in 7 classes,
with 5 fixations in 2 classes. For the remaining of 7 classes, FOD needs more
than 5 fixations to achieve SW's performance. 

MAP-C performs quite well ($84.8\%$ relative performance) even with 1 fixation. The
reason behind this result is the fact that, on average, bounding boxes in the
PASCAL dataset cover a large portion of the images (average bounding box area
normalized by image area is $0.2$) and are located at and around the center
\cite{Tatler2007}. To reduce the effects of these biases about the location of object
placement on the results, we assessed the models with an initial fixation close
to the edge of the image (MAP-E).  When the initial fixation is closer to the
edge of the image,  performance is initially worse than when the initial
fixation is at the center of the image,  The difference in performance diminishes
achieving similar performance with five fixations ($0.2$ difference in mean AP).
Figure  \ref{fig:MAPE_vs_SW} shows how the distribution of AP scores for different object classes for MAP-E improves from 1 fixation to 5 fixations

\subsubsection{Importance of the guidance algorithm}
To assess the importance of guided saccades towards the target we compared
performance of the MAP model against FOD that guides eye movements based on a
random eye movement generator.

Figure \ref{fig:relative_mAP} allows comparisons of the relative performance of the MAP FOD and
those with a random eye movement strategy.  The performance gap between MAP-C,
RAND-C pair and MAP-E,RAND-E pair shows that MAP eye movement strategy is
effective in improving the performance of the system.

\begin{table*}
    \caption{AP scores and relative computational costs of FOD-DPM and DPM on the PASCAL VOC 2007 dataset.}
    \label{table:FOD_with_DPM_results}
\centering
%\scriptsize
\footnotesize
\begin{tabular}{c|c|*{20}{c}|c||c}
%\multicolumn{2}{c|}{} &\rotatebox[origin=c]{50}{\bf aero}& \rotatebox[origin=c]{50}{\bf bike}& \rotatebox[origin=c]{50}{\bf bird}& \rotatebox[origin=c]{50}{\bf boat}& \rotatebox[origin=c]{50}{\bf bottle}& \rotatebox[origin=c]{50}{\bf bus}& \rotatebox[origin=c]{50}{\bf car}& \rotatebox[origin=c]{50}{\bf cat}& \rotatebox[origin=c]{50}{\bf chair}& \rotatebox[origin=c]{50}{\bf cow}& \rotatebox[origin=c]{50}{\bf table}& \rotatebox[origin=c]{50}{\bf dog}& \rotatebox[origin=c]{50}{\bf horse}& \rotatebox[origin=c]{50}{\bf mbike}& \rotatebox[origin=c]{50}{\bf person}& \rotatebox[origin=c]{50}{\bf plant}& \rotatebox[origin=c]{50}{\bf sheep}& \rotatebox[origin=c]{50}{\bf sofa}& \rotatebox[origin=c]{50}{\bf train}& \rotatebox[origin=c]{50}{\bf tv}& \rotatebox[origin=c]{50}{\bf mAP} \\
\multicolumn{2}{c|}{} &\bf aero& \bf bike& \bf bird& \bf boat& \bf bottle& \bf
                   bus& \bf car& \bf cat& \bf chair& \bf cow& \bf table& \bf
                   dog& \bf horse& \bf mbike& \bf person& \bf plant& \bf sheep&
              \bf sofa& \bf train& \bf tv& \bf mAP & \bf Comp. Cost \\
\hline
\multicolumn{2}{c|}{\bf DPM(rel5)} &
33.2 & 60.3 & 10.2 & 16.1 & 27.3 & 54.3 & 58.2 & 23.0 & 20.0 & 24.1 & 26.7 &
12.7 & 58.1 & 48.2 & 43.2 & 12.0 & 21.1 & 36.1 & 46.0 & 43.5 &  33.7 & 100 \\
\hline
\multirow{3}{*}{\bf FOD-DPM} &\bf 1 &
31.0 & 37.1 & 10.0 & 14.3 & 12.9 & 47.1 & 46.7 & 28.0 & 9.3 & 15.5 & 26.2 & 10.7
     & 56.0 & 39.7 & 29.4 & 9.8 & 15.5 & 27.6 & 43.4 & 21.5 &  26.6 & 0.46 \\
 & \bf 5 &
32.3 & 50.0 & 9.8 & 15.2 & 21.8 & 50.0 & 63.0 & 25.9 & 17.1 & 20.5 & 25.4 & 9.7
     & 61.4 & 44.6 & 38.0 & 9.2 & 19.7 & 30.1 & 43.1 & 32.1 &  31.0 & 1.84 \\
 & \bf 9 &
33.2 & 56.6 & 9.9 & 15.6 & 25.3 & 54.6 & 65.3 & 25.3 & 19.8 & 22.0 & 24.9 & 9.4
     & 60.9 & 50.8 & 41.7 & 10.0 & 20.4 & 34.9 & 44.3 & 37.3 &  33.1 & 3.09 \\
     & \bf 13 &
\bf 33.4 & 59.9 & 10.0 & 15.7 & 27.2 & \bf 54.8 & \bf 65.7 & \bf 25.0 & \bf 20.5 & 22.0 & 24.8 & 9.2
     & \bf 62.0 & \bf 51.9 & \bf 44.5 & 10.2 & 20.9 & \bf 36.8 & \bf 46.2 & 40.9
     & \bf 34.1 & 4.16 \\ 
\hline
\end{tabular}

\end{table*}

\subsection{Computational cost} 
The computational complexity of the SW method is easily expressed in terms of
image size. However, this is not the case for our model. 
The computational complexity of FOD is $O(mn)$ where $m$ is the number of fixations
and $n$ is the total number of components, hence templates, on the visual field.  These numbers
do not explicitly depend on the image size; so in this sense, the complexity of
FOD is $O(1)$ in terms of image size. Currently, $m$ is given as an input
parameter but if it were to
be automated, e.g. to achieve a certain detection accuracy, $m$ would implicitly
depend on several factors such as the
difficulty of the object class, the location and size distribution of positive
examples. Targets that are small (relative to the image size) and that are
located far away from the initial fixation location would require more fixations
to get a certain detection accuracy. The number of components, $n$, depends on both the visual field
parameters (number of angle and eccentricity bins which, in our case, are fixed based on the
Freeman-Simoncelli model \cite{freeman_simoncelli:metamers}) and the bounding box
statistics of the target object. These dependencies make it difficult to express
the theoretical complexity in terms of input image size. For this reason, we
compare the computational costs of FOD and SW in a practical framework,
expressed in terms of the total number of operations performed in template evaluations.

In both SW based methods and the FOD, linear template evaluations, i.e. taking
dot-products, is the main time consuming operation. We define the computational
cost of a method based on the total number of template evaluations it executes
(as also done in \cite{vandeSande:iccv2011}). A model may have several templates
with different sizes, so instead of counting each template evaluation as 1
operation,
we take into account the dimensionalities of the templates.  For example, the
cost of evaluating a (6-region)x(8-region) HOG template is counted as 48
operations.

It is straightforward to compute the computational cost (as defined above) of
the SW method. For the FOD, we run the model on a subset of the testing set  and
count the number of operations actually performed. Note that, in order to
compute a detection score, the FOD first performs a feature pooling (based on the
location of the component in the visual field) and then a linear template
evaluation. Since these are both linear operations, we combine them into a
single linear template.  The last column of Table \ref{table:FOD_ap_results}
gives the computational
costs of the SW method and the FOD.  For the FOD the computational cost is
reported as a function of different number of fixations. For ease of
comparison, we normalized the costs so that the SW method  performs 100
operations in total. The results show that FOD is computationally more efficient
than SW. FOD achieves $98.8\%$ of SW's performance at $49.6\%$ of the computational cost of SW. Note that this saving is not directly comparable to that of the
cascaded detection method reported in \cite{felzenswalb:cvpr2010}
because FOD's computational savings
comes about from fewer root filter evaluations, whereas in
\cite{felzenswalb:cvpr2010} a richer model (DPM, root filters and part filters)
is used and the savings are associated to fewer evaluations in the part filters
(i.e., the model applies the root filters at all locations first and
sequentially running other filters on the non-rejected locations).

\begin{figure}
    \centering
    \includegraphics[scale=.45]{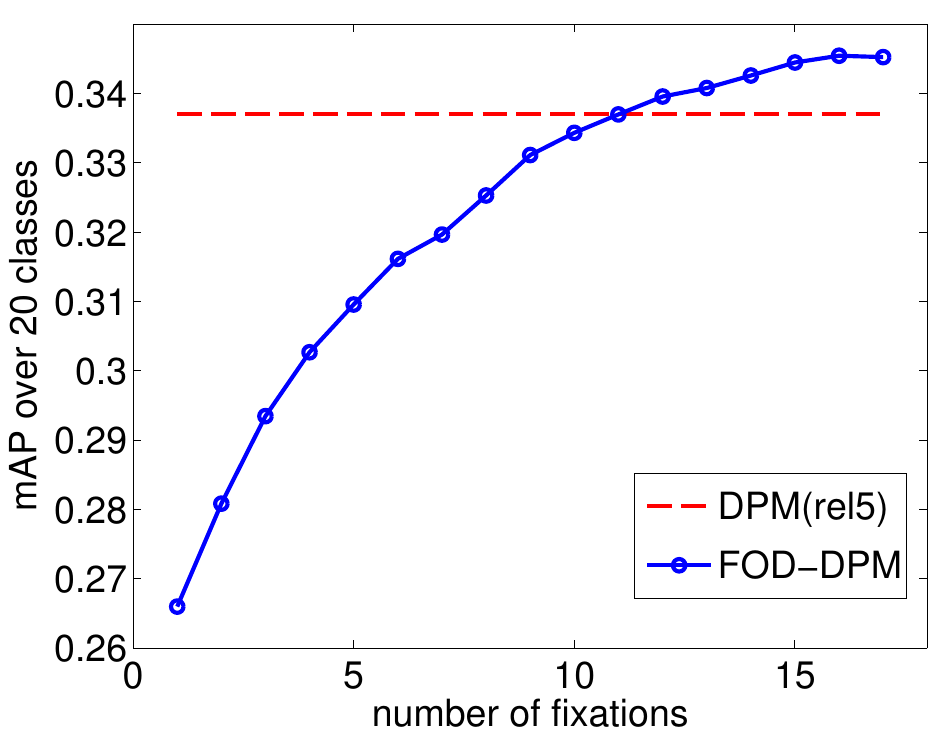}
    \caption{
        \label{fig:DPMvsFODDPM}
        FOD-DPM's performance (mean AP over 20 classes) as a
        function of number of fixations. FOD-DPM achieves DPM's performance at
        11 fixations and exceeds it with more fixations.
    }
\end{figure}

\subsection{Using richer models to increase performance} 
\begin{figure}
    \centering
    \includegraphics[scale=.31]{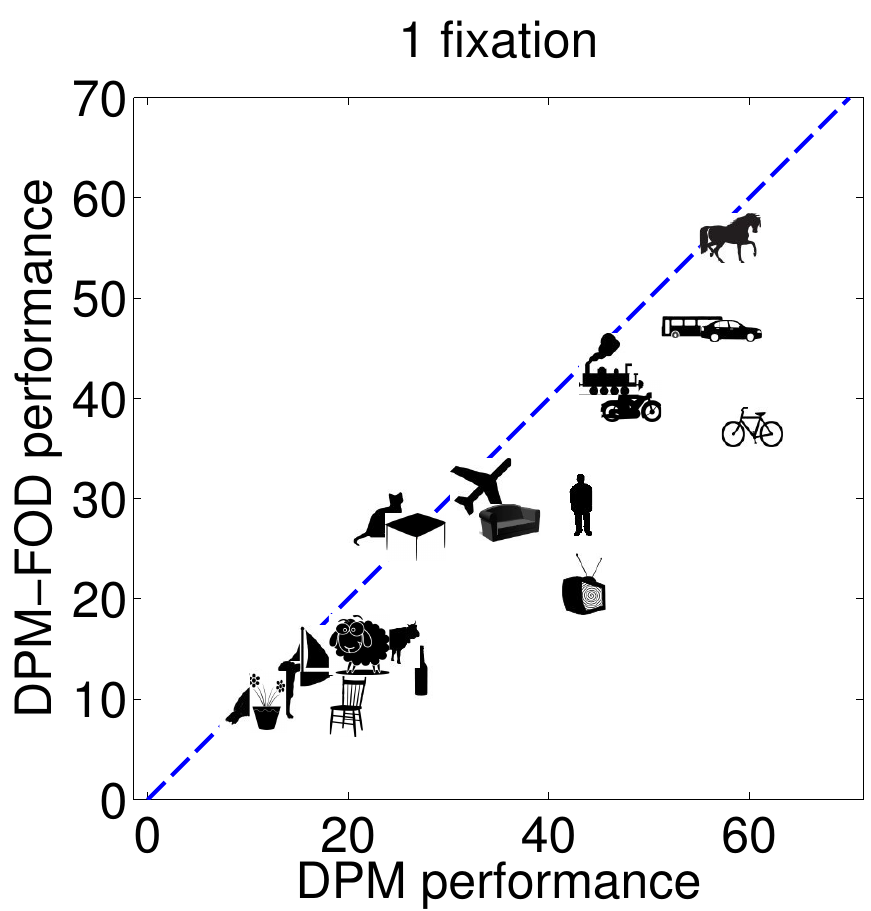}
    \includegraphics[scale=.31]{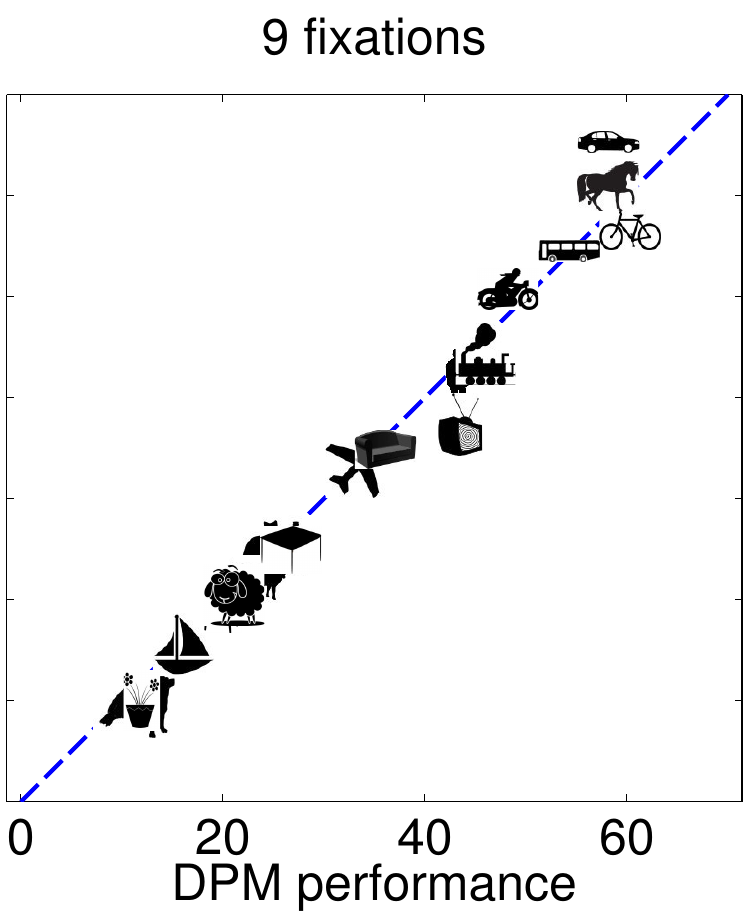}
    \caption{
AP scores achieved by FOD-DPM and DPM per class. 
\label{fig:FODDPM_vs_DPM}
    }
%\end{wrapfigure}
\end{figure}

To directly compare the
computational savings of the FOD model to a cascade-type object detector,
we used a richer and more
expensive detection model at the fovea. This is analogous to the cascaded
detection idea where cheaper detectors are applied first and more expensive
detectors are applied later on the locations not rejected by the cheaper
detectors. To this end, we run our FOD and after each fixation we evaluate the
full DPM detector (root and part filters together) \cite{dpm-voc-release5} only at foveal locations that score above a
threshold which is determined on the training set to achieve high
recall rate ($95\%$). We call this approach ``FOD-DPM cascade" or FOD-DPM for
short. Table \ref{table:FOD_with_DPM_results} and Figure \ref{fig:DPMvsFODDPM} give the
performance result of this approach. FOD-DPM achieves a similar average
performance to that
of  DPM ($98.2\%$ relative performance, $0.6$ AP gap) using 9 fixations and
exceeds DPM's performance starting from 11 fixations. On some
classes (e.g. bus, car, horse), FOD-DPM exceeds DPM's performance probably due
to lesser number of evaluations and reduced false positives; on other cases (e.g.
bike, dog, tv) FOD-DPM underperforms probably due to low recall rate of the
FOD detector for these classes.  Figure \ref{fig:FODDPM_vs_DPM} gives per class
AP scores of FOD-DPM and DPM to demonstrate the improvement from 1 to 9
fixations.

%Comparison of theoretical computational complexities of FOD-DPM and DPM is not
%easily tractable (see the paragraph on ``Computational cost" above), which leaves us
%with practical comparison measures such as wall time. However, neither the wall time
%serves the purpose of comparison because our current MATLAB implementation is
%not as efficient as DPM.
We compare the computational complexities of FOD-DPM and DPM  by their total number of operations  as defined above. For a given object class, DPM model has 3
root filters and 8 6x6 part filters. It is straightforward to calculate the
number of operations performed by DPM as it uses the SW method. 
For
FOD-DPM, the total number of operations is calculated by adding: 1) FOD's
operations and 2) DPM's operations at each high-scoring foveal detection $\bs
b$, one DPM root filter (with the most similar shape as $\bs b$) and 8 parts
evaluated at all locations within the boundaries of this root filter.
Note that we ignore the time for optimal placing of parts in both DPM and
FOD-DPM. Cost of
feature extraction is also not included as the two methods use the same
feature extraction code. We report the computational costs  of FOD-DPM and DPM
in the last column of Table \ref{table:FOD_with_DPM_results}. The costs are
normalized so that DPM's cost is 100 operations. 
Results show  that FOD-DPM drastically reduces the  cost from $100$ to $3.09$
for 9 fixations. Assuming
both methods are implemented equally efficiently, this would translate to an
approximately $32$x speed-up which is better than the $20$x speed-up reported
for a cascaded evaluation of DPM 
\cite{felzenswalb:cvpr2010}. These results demonstrate the effectiveness of our
foveated object detector in guiding the visual search.

\begin{figure*}
    \centering
    \includegraphics[scale=.32]{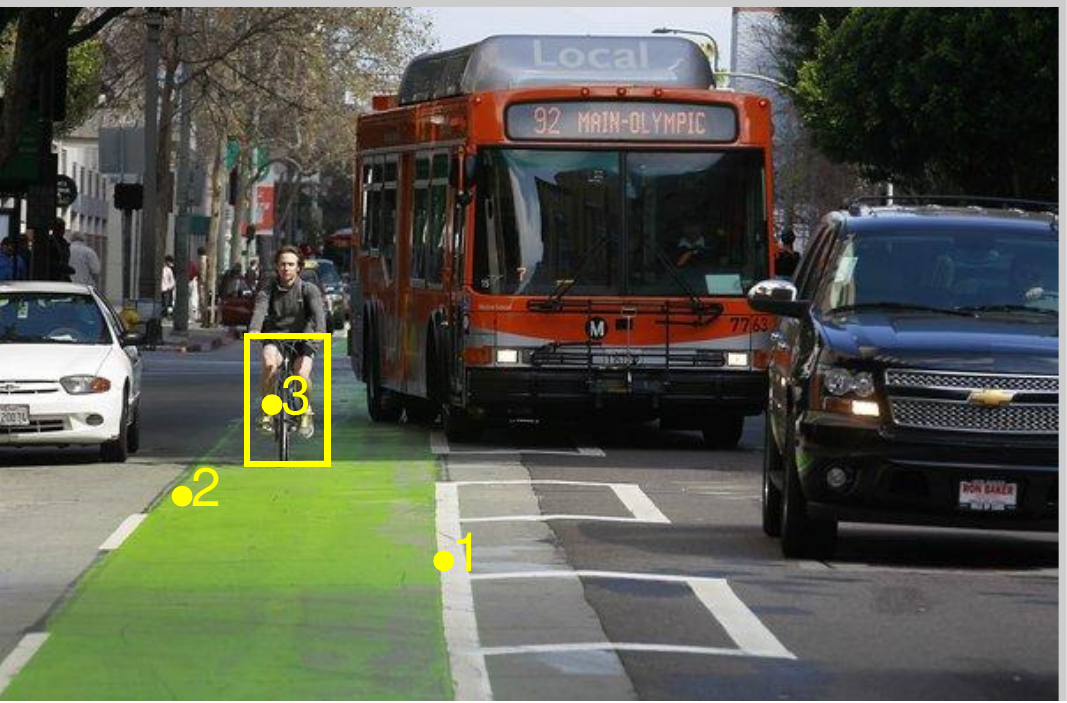}
    \includegraphics[scale=.32]{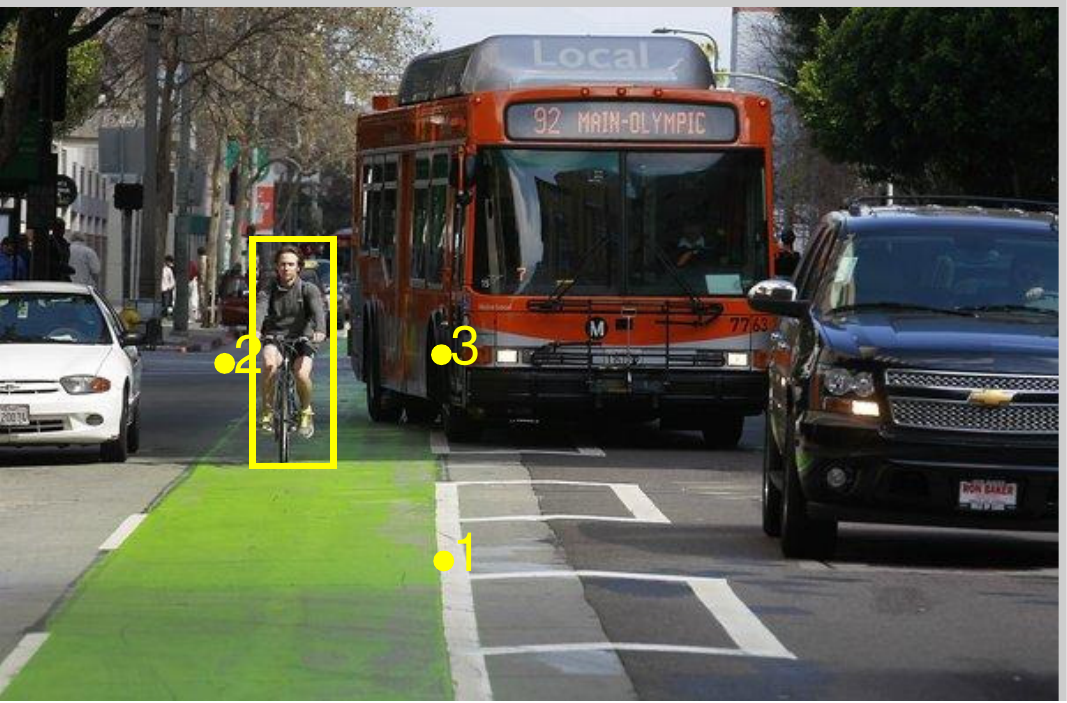} 
    \includegraphics[scale=.32]{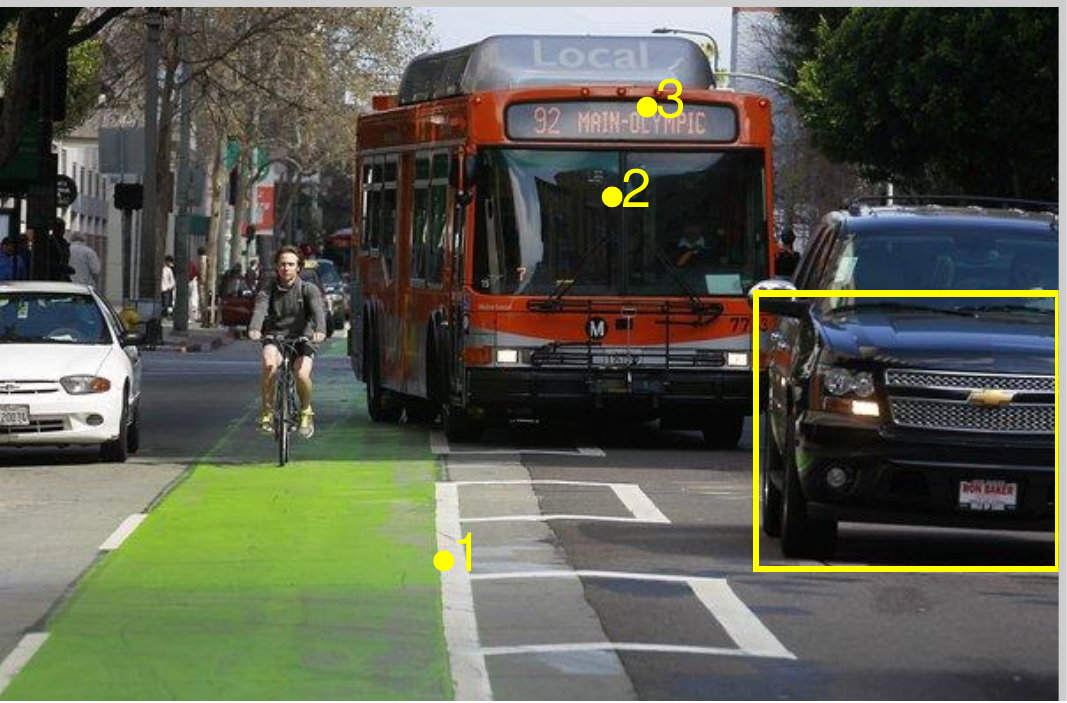}
    \caption{
        Fixation locations and bounding box predictions of FOD for different
        object classes (bicycle, person, and car from left to right) but for the
        same image and initial point of fixation.
        \label{fig:example_detections_on_the_same_image}
    }
\end{figure*}

Finally, in Figure \ref{fig:example_detections_on_the_same_image} we give
sample detections by the FOD system. We ran the trained bicycle, person and car models on
an image outside of the PASCAL datasaet. The models were assigned the same
initial location and we ran them for $3$ fixations. Results show that the each
model fixates at different locations, and these locations  are attracted towards instances of
the target objects being searched. 

\section{Related Work}
\label{sec:related_work}
%There are two prominent models of human eye movements
%in the human vision literature: 1) saccadic targeting or maximum-a-posteriori
%(MAP) model \cite{beutter:2003}, 2) the ideal searcher (IS) model
%\cite{najemnik:nature2005}. Both models compute the posterior probability of a
%target being at a certain location based on observations through multiple
%fixations. MAP model chooses the location with the highest posterior as
%the next fixation point whereas the IS model chooses the location which will
%maximize expected probability of being correct. Both of these models are
%typically used in modeling human eye movements in visual search tasks with
%artificial stimuli which allows rigorous statistical modeling of the task. Our
%object detector implements the MAP model as its basic eye movement strategy and
%works on natural images. 

The sliding window (SW) method is the dominant model of search in object detection. The
complexity of identifying object instances in a given image is $\mathcal{O}(mn)$
where $m$ is the number of locations to be evaluated and $n$ is the
number of object classes to be searched for. Efficient alternatives to sliding
windows can be categorized in two groups: (i) methods aimed at reducing $m$,
(ii) methods aimed at reducing $n$. Since typically $m >> n$, the are a larger
number  efforts
in trying to reduce $m$, however, reducing the contribution of the number of
object classes has recently been receiving increasing interest as search for
hundreds of thousands of object classes has started to be tackled
\cite{dean:cvpr2013}. According
to this categorization, our proposed FOD method falls into the first group as it
is designed to locate object instances by making a set of sequential fixations
where in each fixation only a sparse set of locations are evaluated.

\subsection{Reducing the number of evaluated locations ($m$)}
In efforts to reduce the number of locations to be evaluated, one line of research is the
branch-and-bound methods (\cite{lampert:pami2009,kokkinos:nips2011}) where an
upper bound on the quality function of the detection model is used in a global
branch and bound optimization scheme. Although the authors provide efficiently
computable upper bounds for popular quality functions (e.g. linear template,
bag-of-words, spatial pyramid), it might not be trivial to derive suitable upper
bounds for a custom  quality function. Our method, on the other hand, uses
binary classification detection model and is agnostic to the quality function
used.

Another line of research is the casdaded detection framework
(\cite{viola:ijcv2004, felzenswalb:cvpr2010, lampert:cvpr2010}) where a series
of cheap to expensive tests are done to locate the object. Cascaded detection is
similar to our method in the sense that simple, coarse  and cheap evaluations
are used together with complex, fine and expensive evaluations. However, we
differ with it in that it is essentially a sliding window method with a coarse-to-fine heuristic used
to reduce the number of total evaluations.
%$<<<$Both branch-and-bound and cascaded
%detection methods have questionable biological plausibility, which is one of the
%major aims of our method  DO WE NEED THIS SENTENCE? $>>>$. 
Another
coarse-to-fine search scheme is presented in \cite{pedersoli:cvpr2011} where a
set of low to high resolution templates are used. The method starts by
evaluating the lowest resolution template -- which is essentially a sliding window
operation -- and selecting the high responding locations for further processing
with higher resolution templates. Our method, too, uses a set of varying
resolution templates; however, these templates are evaluated at every fixation
instead of serializing their evaluations with respect to resolution. 

In \cite{vandeSande:iccv2011}, a segmentation based method is proposed to  yield a small
set of locations that are likely to corresponds to objects, which are
subsequently used to guide the search in a selective manner. The
locations are identified in an object class-independent way using an
unsupervised multiscale segmentation approach. Thus, the 
method evaluates the same set of locations regardless of which object class
is being searched for. In contrast, in our method, selection of locations to be
foveated is guided by learned object class templates.

 The method in \cite{alexe:nips2012}, similar to ours,  works like a fixational
 system: at a given time
step, the location to be evaluated next is decided based on previous
observations. However, there are important differences. In \cite{alexe:nips2012}, only a single location is evaluated at a time step
whereas we evaluate all template locations within the visual field at each
fixation. Their method returns only one box as the result whereas our method is
able to output many predictions. 

There are also vector quantization based
methods \cite{kokkinos:eccv2012, sadeghi:nips2013, jegou:pami2011} aiming to reduce the
time required to compute linear template evaluations. These methods to reduce the
contribution of $m$ in $\mathcal{O}(mn)$ are orthogonal to our foveated
approach. Thus, vector quantization approaches can be integrated with the
proposed foveated object detection method. 

\subsection{Reducing the number of evaluations of object classes($n$)}
Works in this group aim to reduce the time complexity contributed by the number
of object classes. The method proposed in \cite{dean:cvpr2013} accelerates the search by replacing the
costly linear convolution by a locality sensitive hashing scheme that works on
non-linearly coded features. Although they evaluate all locations in a given
image, their approach scale constantly over the number of classes, which enables
them to evaluate thousands of object classes in a very short amount of time. 

Another method \cite{song:eccv2012} uses  a  sparse representation of object part
templates, and then uses the basis of this representation to reconstruct
template responses. When the number of object categories is large, sparse
representation serves as a shared dictionary of parts and accelerates the search. 

Another line of research (e.g.
\cite{razavi:cvpr2011,gao:iccv2011,bengio:nips2010}) accelerate
search by constructing classifier hierarchies. These generally work by
pruning unlikely classes while descending the classifier hierarchy.

Importantly, the way the methods in this group accelerate search  is orthogonal
to the savings proposed by using a foveated visual field. Therefore, these
methods are complementary and  can be integrated with our
method to further accelerate search. 

In the context of the references listed in this and the previous sections, our
method of search through fixations using a non-uniform foveated visual field is
novel.

\subsection{Biologically inspired methods}
There have been previous efforts, (e.g. \cite{serre:cvpr2005}), on biologically inspired
object recognition. However, these models do not have a foveated visual field
and thus do not execute eye movements. More recent work has implemented
biologically inspired search methods. In
\cite{elder:ijcv2007}, a fixed, pre-attentive, low-resolution wide-field camera is
combined with  a shiftable, attentive, high-resolution narrow-field
camera, where  the pre-attentive camera generates saccadic targets for the
attentive, high-resolution camera. The fundamental difference between this and
our method is that while their pre-attentive system has the same coarse
resolution everywhere in the visual field, our method, which is a model of the
V1 layer of the visual cortex, has a varying resolution that depends on the
radial distance to the center of the fovea.
There have been previous efforts to create foveated search models with eye
movements \cite{najemnik:nature2005, zhang:2010, Renninger2004, Morvan2012}. Such models have been applied
mostly to detect simple signals in computer generated noise
\cite{najemnik:nature2005, zhang:2010} and used as
benchmarks to compare against human eye movements and performance. 

%Yet none of these models have been
%applied to the problem of object detection in the wild, to standard computer
%vision databases. The models do not
%include a  recent neurobiological implementation of the foveated visual system,
%state of the art object detection algorithms, nor retinotopic location specific
%machine learning.

Other biologically inspired methods include the target acquisition model (TAM)
\cite{zelinsky:nips2006, zelinsky:psych_review2008}, the Infomax model
\cite{infomax_butko:2010} and artificial neural network based models
\cite{larochelle:nips2010, bazzani:icml2011}. TAM is a foveated model and it
uses scale invariant feature transform (SIFT) features \cite{Lowe2004} for representation and utilizes a training set of images to learn
the appearance of the target object. However, it does not include  the variability in object
appearance due to scale and viewpoint, and the evaluation is done by placing the
objects on a uniform background.
%These limitations  make TAM more
%like a proof-of-concept model rather than an object detector that works on fully
%realistic natural images.
The Infomax, on the other hand, can use any previously
trained object detector and works on natural images. They report successful
results on a face detection task. Both TAM and Infomax uses the same template
for all locations in the visual field while our method uses different templates
for different locations. \cite{larochelle:nips2010} was applied to image
categorization and \cite{bazzani:icml2011} to object tracking in videos. 
Critically, none of these models have been tested on standard object detection
datasets nor they have been
compared to a SW approach to evaluate the potential performance loss and
computational savings of modeling a foveated visual field.

% #########################################################################
\section{Conclusions and Discussion}
We present an implementation of a foveated object detector with a recent
neurobiologically plausible model of pooling in the visual periphery and report
the first ever evaluation of a foveated object detection model on a standard
data set in computer vision (PASCAL VOC 2007). 
Our results show that the foveated method achieves
nearly the same
performance as the sliding window method at $49.6\%$ of sliding window's
computational cost. Using a richer model (such as DPM \cite{dpm:pami2010}) to
evaluate high-scoring locations, FOD is able to outperform the DPM  with more computational savings than a state-of-the-art
cascaded detection system \cite{felzenswalb:cvpr2010}. 
These results suggest that using a foveated visual system
offers a promising potential for the development of more efficient object
detectors.
%Also, reducing the computational cost with a slight drop in performance
%might suggest a possible reason for the evolution of foveated visual systems in
%many organisms.
%Regarding the evaluation of object
%detectors for visual search, the PASCAL dataset has issues like center-bias and
%large objects. For the development of alternative search models,
%a new dataset without the mentioned issues seems crucial.
%We believe that foveated object detection seems to be a fruitful
%research direction. For future work, we plan to extend our model by adding the
%non-linear features of V2 and replacing the current generative model based
%machinery with a discriminative model. 
%future work: instead of generative modeling (LDA), train the system using a discriminative system such as SVM

%discussion: biologically-inspired vs biologically plausable

%% use section* for acknowledgement
%\ifCLASSOPTIONcompsoc
  %% The Computer Society usually uses the plural form
  %\section*{Acknowledgments}
%\else
  %% regular IEEE prefers the singular form
  %\section*{Acknowledgment}
%\fi
%Supported by the Institute for Collaborative Biotechnologies through grant
%W911NF-09-0001 from the U.S. Army Research Office. The  content of the
%information does not necessarily reflect the position or the policy of the
%Government, and no official endorsement should be inferred.

\ifCLASSOPTIONcaptionsoff
  \newpage
\fi

\appendices

\section{Approximation of the Bayesian decision}
\label{apx:bayesian_decision}
Derivation for Equation \eqref{eq:FOD_decision}:
\begin{equation}
    \small
    \frac{
    P(y_{\bs b}=1|\bs f_1,  \dots, \bs f_m,I) }
    {P(y_{\bs b}=0|\bs f_1,  \dots, \bs f_m,I) } = 
 \frac{
    P(f_1,  \dots, \bs f_m|y_{\bs b}=1,I) P(y_{\bs b}=1|I)}
    {P(f_1,  \dots, \bs f_m|y_{\bs b}=0,I) P(y_{\bs b}=0|I)}
\end{equation}

\begin{equation}
    \approx \prod_{i=1}^m \frac{P(\bs f_i | y_{\bs b}=1, I)}
    {P(\bs f_i | y_{\bs b}=0 , I)} \frac{P(y_{\bs b}=1|I)}{P(y_{\bs b}=0|I)}
    = \prod_{i=1}^m\frac{P( y_{\bs b}=1| \bs f_i , I)} {P(y_{\bs b}=0 | \bs f_i,
    I)}.
\end{equation}

\section{Detection score after multiple fixations}
Derivation for Equation \eqref{eq:sum_of_scores}:
\label{apx:sum_of_scores}
\begin{equation}
    \mathrm{log}\left(\prod_{i=1}^m \frac{P(y_{\bs b}=1 | \bs f_i, I)}
    {P(y_{\bs b}=0 | \bs f_i, I)} \right)  = 
    \sum_{i=1}^m
    \mathrm{log}\left( \frac{P(y_{\bs b}=1 | \bs f_i, I)}{1-P(y_{\bs b}=1 | \bs f_i, I)}\right)
\end{equation}

using \eqref{eq:score_to_posterior}, we get 
\begin{equation}
    \sum_{i=1}^m
    \mathrm{log}\left( \frac{P(y_{\bs b}=1 | \bs f_i, I)}{1-P(y_{\bs b}=1 | \bs f_i,
    I)}\right) = 
    \sum_{i=1}^m
    \mathrm{log}\left( \frac{\frac{1}{1+e^{-s(I,\bs b, \bs f_i)}}}{1- 
    \frac{1}{1+e^{-s(I,\bs b, \bs f_i)}}}\right)
\end{equation}
\begin{equation}
    = 
    \sum_{i=1}^m
    \mathrm{log}\left( \frac{1}{ e^{-s(I,\bs b, \bs f_i)}}\right)
    =
    \sum_{i=1}^m s(I,\bs b, \bs f_i).
\end{equation}



%\bibliographystyle{IEEEtran}

%{\small
\bibliographystyle{ieee}
\bibliography{references.bib}
\end{document}